\documentclass[10pt,journal,compsoc]{IEEEtran}

\ifCLASSOPTIONcompsoc
  \usepackage[nocompress]{cite}
\else
  \usepackage{cite}
\fi

\usepackage{graphicx}
\ifCLASSINFOpdf
\else
\fi

\usepackage{caption}
\usepackage{subcaption}
\usepackage{graphicx}
\graphicspath{{figures/}}

\usepackage{amssymb}
\usepackage{amsmath} 
\usepackage{amsfonts}
\usepackage{multirow}
\usepackage{booktabs}
\usepackage{arydshln}
\usepackage{cleveref}

\usepackage[dvipsnames]{xcolor}
\newcommand{\C} {black}
\newcommand{\A}[1] {\textcolor{\C}{#1}}

\begin{document}

\title{Are 3D Face Shapes Expressive Enough for Recognising Continuous Emotions and Action Unit Intensities?}

\author{Mani Kumar Tellamekala, Ömer Sümer,  Björn W. Schuller, Elisabeth André, Timo Giesbrecht, Michel Valstar

\IEEEcompsocitemizethanks{

\IEEEcompsocthanksitem Mani Kumar Tellamekala and Michel Valstar are with BlueSkeye AI, UK. This work was done while they were with the Computer Vision Laboratory, School of Computer Science, University of Nottingham, UK.  \protect\\
E-mail:\{mani, michel\}@blueskeye.com

\IEEEcompsocthanksitem Ömer Sümer and Elisabeth André are with the Chair for Human-Centered Artificial Intelligence, University of Augsburg, Germany.  \protect\\
E-mail:\{oemer.suemer, 	andre\}@informatik.uni-augsburg.de

\IEEEcompsocthanksitem Björn W. Schuller is with the Chair of Embedded Intelligence for Health Care \& Wellbeing, University of Augsburg, Germany, and GLAM – the Group on Language, Audio, \& Music, Imperial College London, UK \protect\\
Email: schuller@uni-a.de

\IEEEcompsocthanksitem Timo Giesbrecht is with Unilever R\&D Port Sunlight, UK. \protect\\
Email: timo.giesbrecht@unilever.com
}

}

\markboth{ACCEPTED TO IEEE TRANSACTIONS ON AFFECTIVE COMPUTING}
{}

\IEEEtitleabstractindextext{%
\begin{abstract}
Recognising continuous emotions and action unit (AU) intensities from face videos, requires a spatial and temporal understanding of expression dynamics. Existing works primarily rely on 2D face appearance features to extract such dynamics. This work focuses on a promising alternative based on parametric 3D face alignment models, which disentangle different factors of variation, including expression-induced shape variations. We aim to understand how expressive 3D face shapes are in estimating valence-arousal and AU intensities compared to the state-of-the-art 2D appearance-based models. We benchmark \A{five} recent 3D face models: ExpNet, 3DDFA-V2, \A{RingNet,} DECA, and EMOCA. In valence-arousal estimation, expression features of 3D face models consistently surpassed previous works and yielded an average concordance correlation of .745 and .574 on SEWA and AVEC 2019 CES corpora, respectively. We also study how 3D face shapes performed on AU intensity estimation on BP4D and DISFA datasets, and report that 3D face features were on par with 2D appearance features in recognising AUs 4, 6, 10, 12, and 25, but not the entire set of AUs. To understand this discrepancy, we conduct a correspondence analysis between valence-arousal and AUs, which points out that accurate prediction of valence-arousal may require the knowledge of only a few AUs.
\end{abstract}

}

\maketitle

\IEEEdisplaynontitleabstractindextext

\IEEEpeerreviewmaketitle

\IEEEraisesectionheading{\section{Introduction}\label{sec:introduction}}
\label{sec:intro}

\IEEEPARstart{F}{acial} expressions are important social signals produced through coordinated movements of facial muscles along spatio-temporal dimensions. Automatic recognition of facial expressions from video data is a fundamental task in Affective Computing with a wide range of applications, including but not limited to psychotherapy and well-being~\cite{Girard:2013}, educational analytics~\cite{wu2016review}, naturalistic human-computer~\cite{Dornaika:2009}, and human-robot interactions~\cite{Faria:2017}. The problem of automated facial expressive behaviour analysis has been extensively studied in the last two decades~\cite{valstar2006fully, valstar2015fera, ringeval2019avec, kollias2020analysing}. The two most common video-based facial expression analysis approaches are based on Russell's circumplex model of dimensional emotions~\cite{russell1980circumplex} and Facial Action Coding System (FACS)~\cite{ekman1977facial}. The circumplex model represents emotions in a continuous space composed of two orthogonal axes, namely valence and arousal dimensions. In contrast, FACS encodes the movements of different facial muscle groups by defining the occurrence and intensity values of their corresponding Action Units (AUs). 
 
\begin{figure}

\includegraphics[width=1.05\linewidth]{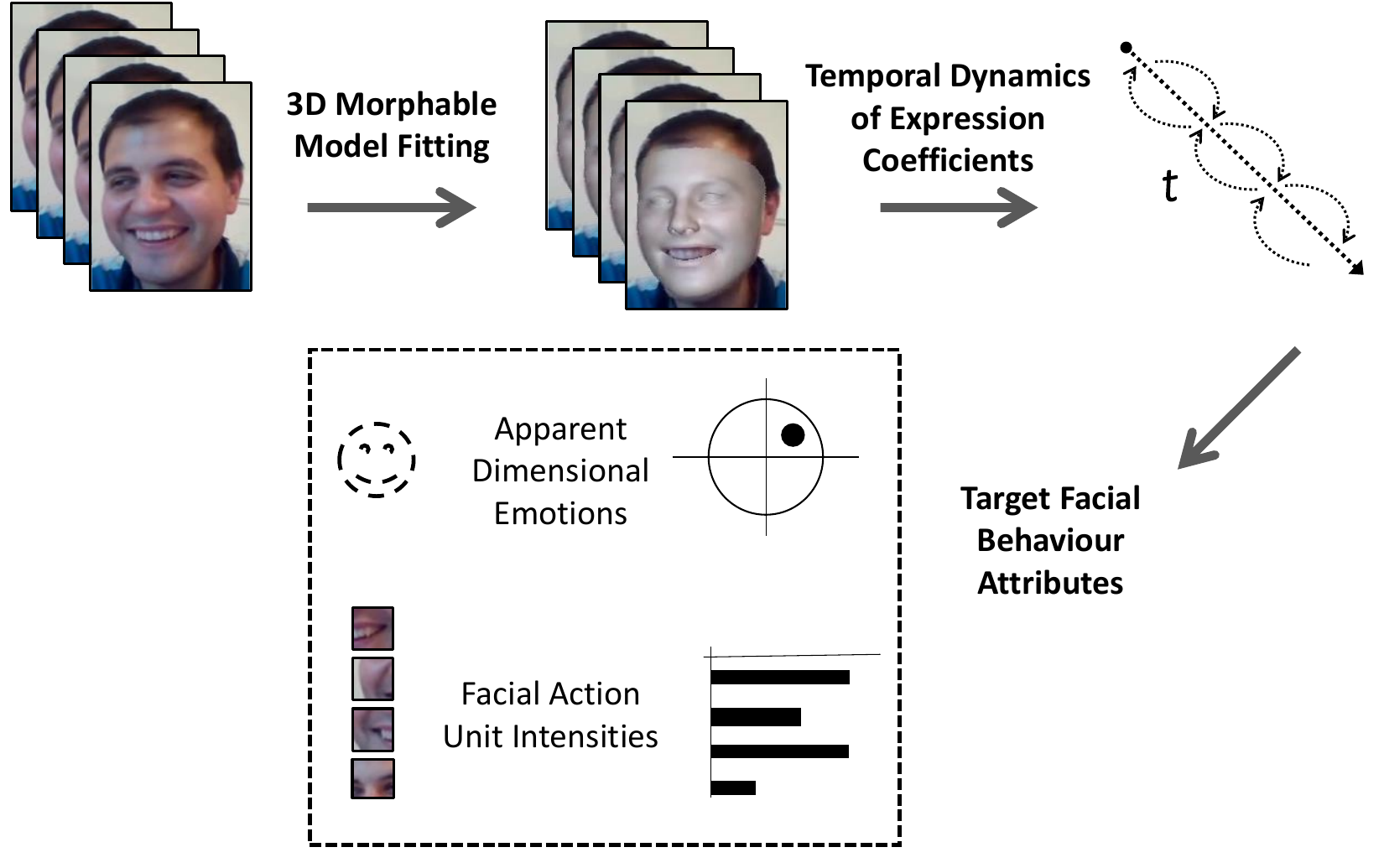}
   \caption{Recognising continuous dimensional emotions and facial action unit intensities from the temporal dynamics of of 3D Morphable Models' expression coefficients}
\label{fig:trailer}
\end{figure}
 
A common challenge encountered in video-based facial expression analysis in naturalistic conditions is to disentangle expression-induced facial variations from a wide range of other factors of variation in a given 2D face image sequence. Expression-irrelevant facial variations typically include head pose changes, facial geometry that contains identity information, or other fine-scale details such as identity-specific face wrinkles, etc. In the era of deep representation learning, most state-of-the-art methods depend on end-to-end learning from 2D face appearances. Such 2D appearance-based expression features achieved impressive performance on both valence-arousal and AU intensity estimation tasks~\cite{toisoul2021estimation,ntinou2021transfer,kollias2020exploiting}. However, they heavily rely on the manual annotations of emotions or AU intensities for vast amounts of visual data to extract facial expression features and their temporal dynamics. In contrast, analysis-by-synthesis methods such as 3D Morphable Models (3DMM)~\cite{Blanz:1999} of faces offer an interesting alternative to distil expression-induced facial shape variations in a more principled approach. Such analysis-by-synthesis methods, most importantly, do not need the labels of emotions or AUs to extract expression features from 2D face images. 

Several parametric 3D face alignment models~\cite{Zhu:2019:3DDFA_v1, Guo:2020:3DDFA_v2, RingNet:CVPR:2019, Feng:2021:DECA,Danecek:2022:EMOCA} based on 3DMM formulation achieved significant improvements in recent years by leveraging advancements in data-driven representation learning. Some recent works on 3D face alignment even attempted to reconstruct facial expressions with high fidelity~\cite{Chang:2018:ExpNet,Feng:2021:DECA, Danecek:2022:EMOCA}. Despite such improvements in 3D face alignment methods, the idea of utilising 3DMM expression information for video-based facial expression analysis received limited attention compared to 2D appearance-based approaches. On that regard, we pose the following questions in this work: 
\begin{itemize}
    \item Are 3D face shapes expressive enough to estimate AU intensities as well as dimensional emotions (valence-arousal) from face video data?
    \item Where do 3D face shape expression features stand w.r.t.\ 2D face appearance features that are directly learned for estimating AU intensities and emotions in an end-to-end fashion? 
\end{itemize}

To answer these questions, as Fig.~\ref{fig:trailer} illustrates, we train AU intensity estimation and dimensional emotion recognition models based on the temporal dynamics of 3D facial expressions. We extensively evaluate the quality of 3DMM based expression features on the datasets of valence-arousal estimation (SEWA~\cite{kossaifi2019sewa}, AVEC 2019 CES~\cite{ringeval2019avec}) and AU intensity estimation (BP4D~\cite{zhang2014bp4d} and DISFA~\cite{mavadati2013disfa}). We apply a simple bi-directional \A{Gated Recurrent Unit (GRU)} network to model the temporal dynamics of 3DMM expression features extracted from \A{five} dense 3D face alignment models: ExpNet~\cite{Chang:2018:ExpNet}, 3DDFA-V2~\cite{Guo:2020:3DDFA_v2}, \A{RingNet~\cite{RingNet:CVPR:2019},} DECA~\cite{Feng:2021:DECA}, and EMOCA~\cite{Danecek:2022:EMOCA}. We compare the recognition performance of different 3D face shape models with the 2D face appearance baselines and models that currently have state-of-the-art performance on both tasks.

Our experimental analysis shows that in the case of continuous emotion recognition 3D face expression features outperform the existing benchmarks as well as the 2D appearance baselines evaluated in this work. However, on the task of AU intensity prediction 3D face shape models perform poorly compared to the existing state-of-the-art benchmarks based on 2D appearance features. Further, we conduct a correspondence analysis between different AUs and valence-arousal dimensions to explain the performance discrepancy of 3D face models between emotion recognition and AU intensity prediction tasks. Thus, this work comprehensively illustrates the current state of the 3D face shape expression features in terms of their ability to model continuous facial expression dynamics. 


\section{Background and Related Work}
\label{sec:rel_work}

As the main focus of this work is the analysis of 3D face shape models for facial expression analysis, we review the literature on 3D morphable models of faces and expression analysis tasks tackled by using 3D face features. Here, our particular interest is facial geometry-aware approaches in two video-based facial expression analysis tasks: valence-arousal estimation and AU intensity estimation.

\subsection{3D Morphable Models of Faces}
Estimating 3D shape models from 2D measurements is a fundamental problem in computer vision. Mainly focusing on face analysis, Blanz and Vetter~\cite{Blanz:1999} initially addressed this problem and proposed a 3D Morphable Model (3DMM) to generate 3D face shape and appearance. 3DMM can be considered a representation of facial shape and texture, disentangling them from other factors of variation. Fundamentally, 3DMM is a statistical model learned from dense one-to-one correspondences of representative 3D shapes and 2D appearance data. The original work performed face registration from unregistered 3D scenes using gradient-based optical flow and created a 3D face model, learning an optimisation problem to synthesise 2D appearance from a linear combination of 3D shapes using PCA decomposition. We refer the interested readers to \cite{Egger:2020} for a detailed review of 3D morphable face models.

The main factors of variation in a 3DMM are shape and texture. The original formulation can be given as follows:
\begin{equation}
 S = \bar{S} + A_{s}\alpha_{s} + A_{t}\alpha_{t},   
 \label{eq:3dmm}
\end{equation}
where $S$ is a 3D face, $\bar{S}$ is the mean 3D shape, $A_{shape},A_{texture}$ are shape and texture bases, and $\alpha_{shape},\alpha_{texture}$ are their parameters. After the 3D face is reconstructed with this model, it is projected back to the image plane using a scale orthographic model:
\begin{equation}
    V_{2d}(\mathbf{p})=f * \mathbf{Pr} * \mathbf{R} * (\bar{\mathbf{S}}+ \mathbf{A}_{s}\mathbf{\alpha}_{s} + \mathbf{A}_{t}\mathbf{\alpha}_{t}  ) + t_{2d},
    \label{eq:3dmm_proje}
\end{equation}
where $V_{2d}(\mathbf{p})$ generates the 2D locations of 3D model vertices. The head pose information comes from the scale factor $f$, orthographic projection matrix, $\mathbf{Pr}$ and rotation matrix, $ \mathbf{R}$. $\mathbf{\alpha}_s$ contains the identity-related and expression-induced shape information, whereas texture details captured by $\mathbf{\alpha}_t$ correspond to variations that may also contain expression information. However, the original and many following 3DMM approaches focused on reconstructing 3D face shapes per identity and neglected expression variations specifically in their optimisation. \\

\noindent \textbf{Modelling Facial Expressions in 3DMMs. }
Looking into the methods that incorporate facial expression information in 3DMMs, 3DFFA~\cite{Zhu:2019:3DDFA_v1} optimised the bases on 3D models of faces with various expressions in the FaceWarehouse database~\cite{Cao:2014:FaceWarehouse} containing 3D scans of 150 people of diverse ages and ethnic backgrounds. Subsequently, Guo et al.~\cite{Guo:2020:3DDFA_v2} improved the optimisation by computing Vertex Distance Cost (VDC) and Weighted Parameter Distance Cost (WPDC) and by using more compact backbone regressors; however, their expression modelling approach remained the same. 

Similar to 3DDFA~\cite{Zhu:2019:3DDFA_v1}, Chang et al.~\cite{Chang:2018:ExpNet} proposed a landmark-free approach. They estimated 3DMM shape and pose parameters using CNN-based models. By leveraging the identity information and assuming that the shape parameters of a person's different images will remain the same, they acquired the expression deformation using Gauss-Newton optimisation. Subsequently, they extracted expression codes on large-scale face datasets and regressed them with ResNet-101 deep network architecture.

Recently, Li et al.\ (FLAME model \cite{Li:2017:FLAME}) used a large amount of training data to capture intrinsic shape deformations and the deformations related to pose changes, to some extent, modelling facial muscles' activation and respective  expressions. In order to decouple expressions from pose variations, they estimated pose coefficients, applied an inverse transform, and normalised pose to reduce its effect on expression parameters. \A{Sanyal et al.~(RingNet, \cite{RingNet:CVPR:2019}) learned a mapping from RGB images to 3D FLAME model parameters by adding a geometric constraint. They incorporated a shape consistency loss that encourages the same shape for the same persons' face images and different shapes for different persons and a 2D feature loss that predicts 2D facial landmarks by projecting corresponding 3D points from the FLAME template. RingNet used more face scans and additional geometric losses but lacks any explicit emotion terms in the optimization.} Feng et al.~(DECA,\cite{Feng:2021:DECA}), by building on top of the FLAME model, disentangled static and dynamic facial details \A{in unconstrained} images. After reconstructing the coarse shape, they swap person-specific details and jaw pose parameters between different images of the same person and disentangle them from expression.

\A{We should note the initial line of work in 3DMM fitting is based on analysis-by-synthesis optimisation~\cite{blanz2003face,Romdhani:2005,Garrido:2013,Ferrari:2017,Ferrari:2021}. Most of these approaches are sensitive to the initial conditions in their optimisation. They performed well in constrained face capture settings but are susceptible to operating conditions and image quality. Another significant difference is deep learning-based regression methods can capture shape and texture better by learning an encoder representation from large-scale labelled and unlabeled facial data. Furthermore, optimisation-based methods proposed additive and multiplicative methods (similar to Eq.~\ref{eq:3dmm}). However, expressions are highly complex and entangled with other facial traits. Thus, nonlinear expression modelling enhances the quality of captured expressions, as in the FLAME model that combines articulated jaw and eyeballs and linear expression blend shapes. Furthermore, unlike the optimisation-based approaches, deep learning-based methods can also  easily incorporate auxiliary tasks, for instance, emotion recognition in EMOCA~\cite{
Danecek:2022:EMOCA} to improve the expression features of 3D faces.}

\subsection{3D Face Models in Expression Analysis}

\A{This work mainly focuses on using 3D face models in video-based expression analysis tasks, namely, valence-arousal and facial action unit intensity estimation. As alternatives to the standard 2D appearance-based features used in expression analysis models (e.g.~\cite{kollias2017recognition,kumar2018consistent,kollias2018aff,sanchez2018joint}), several previous works explored the use of shape features derived from 3D face models. Such commonly used shape features include landmark locations predicted in 3D face alignment models (e.g.~\cite{fabiano2019deformable}), their displacement vectors (e.g.~\cite{pei2020monocular}), and different parametric representations of 3D face shapes (e.g., ~\cite{chen20153d,koujan2020real}), etc. This section presents a brief overview of the notable works that explored the application of 3D face features in recognising discrete emotions, continuous emotions, and facial action units.}

\A{Most existing methods using 3D facial features for expression analysis focus on discrete emotion recognition. By tracking non-rigid deformations in 3D face surfaces, early works such as Wen and Huang~\cite{wen2003capturing} attempted to classify four discrete emotions (anger, disgust, fear, and sadness) on the selected images of CMU Cohn-Kanade expression database that is limited in terms of subjects and also environmental conditions. In later works, the focus shifted to recognising discrete emotions from 3D face scans through 3DMM parameters. 3DMM methods were initially based on optimisation and inverse graphics, and they aimed at an analysis-by-synthesis mapping between 2D and 3D observations. They mostly used expressive faces and discrete facial expression recognition to validate their 3D model fitting. However, most state-of-the-art 3DMM works use regression methods and apply deep learning on a large number of unconstrained images to gain robustness. Learning facial expressions is either a separate learning task using 3DMM representations or a loss term when optimising deep learning models to regress 3DMM parameters.}

\A{Bejaoui et al.~\cite{bejaoui2017fully} used the face mesh generated by a 3DMM model and combined it with an appearance feature, Local Binary Patterns (LBP), to represent both geometry and appearance. Subsequently, by training an SVM classifier on those features, they reported discrete emotion recognition results on the Bosphorus database~\cite{savran2008bosphorus}. Ferrari et al.~\cite{Ferrari:2017} proposed a dictionary learning-based method for 3DMM fitting and also validated the expressiveness of their approach on Extended Cohn-Kanade (CK+)~\cite{Lucey:2010} and the Facial Expression Recognition and Analysis (FERA)~\cite{valstar2012meta} for discrete emotion recognition and facial action unit detection tasks, respectively. Even though these were the first works that classified discrete emotions from the 3DMM expression parameters, they were based on analysis-by-synthesis 3DMM methods. Their performance was limited, and they were not evaluated in continuous expression tasks.}

\A{There are more recent, deep learning-based examples that used 3DMM features. For instance, Shi et al.~\cite{shi2020pose} trained an encoder-decoder architecture to reconstruct 3DMM expression parameters (not ground truths, predictions of another 3DMM estimator) and perform facial expression classification simultaneously. Their approach achieved facial expression accuracy of 75.63\% and 70.20\% on CK+ and OULU-CASIA databases. Among the learning-based 3DMM methods, ExpNet~\cite{Chang:2018:ExpNet} regressed 3DMM shape parameters using a deep neural network, projected 3D shape to 2D points of the image and estimated expression coefficients using standard Gauss-Newton optimisation. They subsequently used the expression coefficients and a simple kNN classifier to classify discrete facial expressions on the CK+ and the Emotion Recognition in the Wild Challenge (EmotiW-17) dataset. EMOCA~\cite{Danecek:2022:EMOCA} is the current state-of-the-art approach, and it also uses emotion information during training. Building on DECA~\cite{Feng:2021:DECA}, EMOCA learns to minimise a perceptual emotion consistency loss between the emotion features of input images and those of rendered ones.}


\A{DeepExp3D~\cite{koujan2020real} is another recent approach that regresses 3DMM expression parameters independent of a person's identity. They reported discrete emotion recognition performance on static face image datasets like CK+, CFEE, 
etc, where all of them are posed and captured under controlled environments. This line of work showed enough evidence about emotion information learned by 3DMM parameters; however, a comprehensive analysis of 3D face models on real-world tasks, especially on continuous emotion estimation, is largely lacking.}

\A{Pei et al.~\cite{pei2020monocular} and Chen et al.~\cite{chen20153d} are two notable approaches among continuous emotion recognition methods based on 3D face features. Pei et al.~\cite{pei2020monocular} proposed to learn an extended 3DMM model that provides spatiotemporal features for valence-arousal estimation.  Though this method showed better results than different baseline models using CNN-based appearance features, its evaluation did not include any state-of-the-art 3DMM models for comparison.}

\A{In Chen et al.~\cite{chen20153d}, a novel random forest-based joint regression model was proposed for recovering 3D face shapes and estimating valence-arousal values from video data. On a relatively small-scale dataset (AVEC 2012~\cite{schuller2012avec}) composed of videos recorded in controlled settings, this method demonstrated promising emotion recognition performance (Pearson's correlation coefficients of 0.45 and 0.56 for valence and arousal, respectively). However, its performance benchmarking, similar to Pei et al.~\cite{pei2020monocular}, did not consider other 3DMM models as baselines. Thus, it is not clear how well this method performs when compared with different 3DMM models and 2D appearance-based features on video-based valence-arousal estimation, particularly on recently developed large-scale in-the-wild emotion datasets. Though EMOCA partially addressed this problem by benchmarking different 3DMM models on continuous emotion estimation, its evaluation was done on only static face image datasets such as AffectNet~\cite{mollahosseini2017affectnet}. Our work aims to fill this gap by extensively benchmarking five state-of-the-art 3DMM models' valence-arousal estimation performance on two in-the-wild video datasets, SEWA~\cite{kossaifi2019sewa} and AVEC 2019~\cite{ringeval2019avec}.}

\A{Unlike the emotion recognition tasks, not much attention has been paid  so far to AU intensity estimation from videos using 3D face models. Ferrari et al.~\cite{Ferrari:2017} and Ariano et al.~\cite{ariano2021action} studied the use of 3DMM coefficients in detecting AU occurrences in static face images. These works showed that 3DMM coefficients coupled with SVM classifiers can achieve comparable AU detection results to the models using appearance features. However, it is important to note that in the case of AU intensity estimation facial features need to be more fine-grained than the features required for AU detection. Furthermore, it is not known how well the temporal dynamics of 3DMM coefficients capture continuous-valued AU intensities in videos. Aiming to answer these questions, in this work we evaluate the AU intensity estimation performance of state-of-the-art 3DMM models on two benchmark video datasets (BP4D and DISFA).}

\A{In summary, while all the aforementioned existing works partially demonstrated the efficacy of 3D facial features in expression analysis tasks in general, it is not comprehensively understood how well different 3D facial features perform on in-the-wild video data, in comparison with 2D appearance features based on deep representation learning. To this end, we extensively evaluate five state-of-the-art 3D face models on different benchmark datasets of facial emotion and action unit analysis.}

\section{3D Shape vs.\ 2D Appearance Features for Continuous Facial Expression Analysis}
\label{sec:method_exp}

The face is essentially a 3D volumetric surface that undergoes rigid (e.g., head pose changes) and non-rigid (e.g., talking and raising eyebrows) deformations. Capturing such non-rigid deformations that correspond to emotional expressions along spatio-temporal dimensions is at the core of video-based facial expressive behaviour analysis. Here, our goal is to comprehensively compare and analyse the performance of standard 2D CNN-based facial appearance features learned using task-specific target labels and expression-related facial features derived from dense 3D face alignment models. To this end, we model the temporal dynamics of 3D shape-based features and 2D appearance-based features extracted from face image sequences for learning expression analysis tasks. In particular, we consider time-continuous dimensional emotion (valence-arousal) recognition and action unit (AU) intensity estimation as representative tasks for video-based facial expression analysis. In this comparison, it is worth noting that expression features in 3D face models are learned with the objective of accurate shape reconstruction, whereas 2D face appearance features are directly optimised to predict task-specific target labels (valence-arousal or AU intensities).


\subsection{Expression Embeddings from 3D Face Shapes} For extracting expression-specific 3D face shape features, we consider 3D Morphable Models (3DMM)\cite{Blanz:1999} of faces, for they offer a principled approach to factorise the facial expression information. In the standard linear representation of 3DMM used in face alignment, as shown in Eq.~\ref{eq:3dmm}, the shape component can be further decomposed as follows:
\begin{equation}
    \textbf{A}_{s} \textbf{$\alpha$}_{s} = \textbf{A}_{id} \textbf{$\alpha$}_{id}+\textbf{A}_{ex}\textbf{$\alpha$}_{ex}, 
    \label{eq:3dmm_shape}
\end{equation}
where $\textbf{A}_{id}$ and $\textbf{A}_{ex}$ are the basis matrices of shape identity and expressions respectively, and $\textbf{$\alpha$}_{id}$ and $\textbf{$\alpha$}_{ex}$ are their corresponding coefficient vectors. Here, we refer to the coefficient vectors $\textbf{$\alpha$}_{ex}$ as expression embeddings.

Given a 2D face image as input, to extract its expression embedding from its 3D face shape, we consider \A{five} different approaches that learn 3DMM  parameters: ExpNet~\cite{Chang:2018:ExpNet}, 3DDFA-V2~\cite{Guo:2020:3DDFA_v2}, \A{RingNet~\cite{RingNet:CVPR:2019},} DECA~\cite{Feng:2021:DECA}, and EMOCA~\cite{Danecek:2022:EMOCA}.  The criteria for selecting these \A{five} models are as follows: EMOCA~\cite{Danecek:2022:EMOCA} is the current state-of-the-art in capturing 3D facial expressions, building on the ability of DECA~\cite{Feng:2021:DECA} formulation in modelling detailed facial expressions. DECA develops this ability by adopting a consistency loss to effectively disentangle details specific to a person from wrinkles induced by expressions. Next, we choose 3DDFA-V2~\cite{Guo:2020:3DDFA_v2} since its global shape reconstruction error is very close to that of DECA~\cite{Feng:2021:DECA}.  \A{In contrast to the aforementioned models, which use 2D facial geometry only as a regularisation constraint in minimising 3D reconstruction loss, RingNet~\cite{RingNet:CVPR:2019} heavily relies on 2D landmarks for supervision by minimising a novel shape consistency loss.} The last model that we evaluate here is ExpNet~\cite{Chang:2018:ExpNet}, which directly regresses expression coefficients inferred using 3DDFA~\cite{zhu2016face}, a predecessor to 3DDFA-V2. 

Though all these \A{five} models output similar expression embedding vectors ($\alpha_{ex}$), their dimensionality varies from model to model: 29\,D for ExpNet (same as in the original 3DDFA), 10\,D for 3DDFA-V2, and 50\,D for \A{RingNet}, DECA and EMOCA. It is important to note that the fidelity of facial expressions captured by these models does depend not only on the expression embedding dimensionality but also on various other factors such as their corresponding CNN backbone complexity and optimisation procedures followed during their training, etc.




\begin{figure}
\begin{center}
\includegraphics[width=0.99\linewidth]{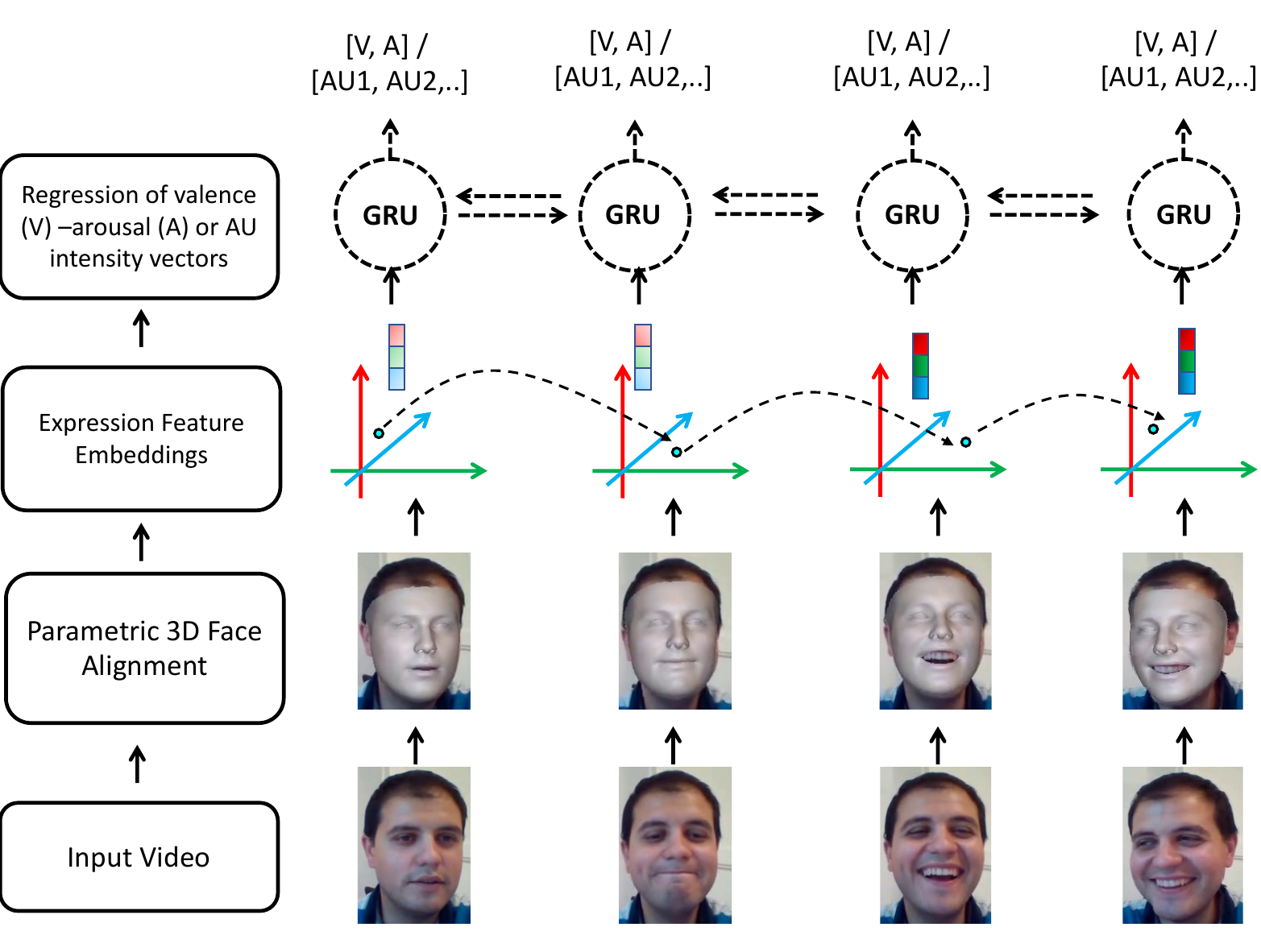}
\end{center}
   \caption{Modelling temporal dynamics of 3DMM expression coefficients using bidirectional GRUs for video-based dimensional emotion recognition and AU intensity estimation}
\label{fig:temp_model}
\end{figure}

\subsection{Expression Embeddings from 2D Face Images} 

We use end-to-end learning models based on the standard CNN + RNN architectures as 2D appearance baselines. For this purpose, we adopt two strong CNN backbones that are extensively used for end-to-end facial feature learning in several recent works~\cite{kollias2018aff,kumar2018consistent,toisoul2021estimation,tellamekala2022modelling}.   

\textbf{ResNet-50}~\cite{he2016deep}, particularly a version of it pre-trained on the VGG-Face database~\cite{parkhi2015deep}, is a commonly used CNN backbone for feature extraction from face images for emotion recognition~\cite{kollias2017recognition,kumar2018consistent, deng2020mimamo}. In implementing this  backbone, we flatten the output feature maps of its last convolutional layer into 2056-dimensional vectors, which we further map to 512-dimensional features using a fully connected layer. Considering the relatively small-scale training datasets available for AU intensity estimation tasks, we use ResNet-18 model, following existing works~\cite{ntinou2021transfer,sanchez2021affective,tellamekala2022modelling}.

\textbf{EmoFAN}~\cite{toisoul2021estimation,sanchez2021affective} is designed for facial feature extraction using only convolution layers to make the model more efficient in terms of the number of trainable parameters. A pre-trained variant of this backbone on 2D face alignment tasks is found to be very effective for transfer learning~\cite{toisoul2021estimation,ntinou2021transfer}. To extract facial features with better generalisation capacity, we use a variant of this CNN backbone pre-trained on image-based emotion recognition using the AffectNet dataset~\cite{mollahosseini2017affectnet}, in addition to the 2D face alignment task. Following prior works~\cite{toisoul2021estimation, sanchez2021affective}, we use this backbone to extract 512-dimensional facial embedding vectors.

\subsection{Temporal Dynamics of Expression Embeddings} Fig.~\ref{fig:temp_model} illustrates the steps that we follow for video-based expression analysis tasks. Modelling temporal dynamics of frame-wise expression features in a video is critical for dimensional emotion recognition and AU intensity estimation tasks. For this purpose, we use a simple bidirectional GRU network with two hidden layers of 128 dimensions. For a fair comparison, we use the same temporal network for the expression embeddings from the 3D shape and 2D appearance models. Note that the dimensionality of the input embeddings varies across the different models. On top of the last layer of GRU block output, there is a single fully connected layer to map the per-frame hidden state vector to the final output vector of dimensional emotions or AU intensities. The output is two-dimensional in valence-arousal prediction models, whereas it differs in the number of action units in AU intensity estimation models (5-dimensional in the BP4D dataset and 12-dimensional in the DISFA dataset).

\subsection{Datasets} 
\noindent \textbf{Dimensional Emotion Recognition.} For video-based valence and arousal estimation, we use two large-scale video datasets: SEWA~\cite{kossaifi2019sewa} and the AVEC'19 Cross-cultural Emotion Sub-Challenge (CES) Corpus~\cite{ringeval2019avec}. 

\textbf{SEWA} dataset was collected during computer-based naturalistic dyadic interactions and contains 538 face videos of 398 subjects from 6 different cultures. Each video is annotated with per-frame continuous-valued valence and arousal annotations in the range of -1 to 1 at 50 frames per second (FPS). The numbers of videos used for training, validation, and testing\footnote{The details of the train, validation, and test partitions were kindly provided by the database owners.} are 431, 53, and 53, respectively, with the duration in the range of 10\,s to 30\,s. 

\textbf{AVEC'19 CES Corpus} is a multimodal in-the-wild affect recognition dataset captured in cross-cultural settings and consists of German, Hungarian, and Chinese subjects. All videos in this dataset are also annotated with continually varying valence and arousal ratings at 50 FPS in the range [-1,1], the same as in the case of SEWA. It provides 64 videos for training, and 32 videos for validation, with a total duration of roughly 160 minutes and 65 minutes, respectively. We report the results on its validation set since the test set labels are not publicly available. \\ 

\noindent \textbf{Action Unit Intensity Estimation.} We use two video-based AU intensity labelled datasets: DISFA~\cite{mavadati2013disfa} and BP4D~\cite{zhang2014bp4d}.

\textbf{DISFA} has 27 videos of 27 subjects; each video contains approximately 4844 frames annotated with the intensity values of 12 AUs. As there are no predefined training, validation and test partitions, a subject-independent 3-fold cross-validation is a commonly used evaluation protocol for this dataset. To compare with the state-of-the-art results on DISFA, following the existing works (e.g. ~\cite{ntinou2021transfer,sanchez2021affective,tellamekala2022modelling}), we also perform the 3-fold cross-validation; each fold containing 18 videos for training and 9 videos for evaluation. 

\textbf{BP4D} has 487 videos of 41 subjects, containing approximately 140,000 frames annotated with the intensity values of 5 AUs. It was the main corpus of the FERA 2015 challenge~\cite{valstar2015fera}. We use the same training (168 videos), validation (160 videos), and test (159 videos) sets that were originally used by the FERA 2015 challenge participants~\cite{valstar2015fera}.  

\subsection{Evaluation Metrics}


\noindent{\textbf{Dimensional Emotion Recognition}} performance is measured using Lin's Concordance Correlation Coefficient (CCC)~\cite{lawrence1989concordance}, which computes the agreement between target emotion labels $e^*$ and their predicted values $e^o$ as,
\begin{equation}
    CCC = \frac{\rho_{e^*e^o} . \sigma_{e^*} . \sigma_{e^o}}{{(\mu_{e^*}-\mu_{e^o})}^2 + {\sigma_{e^*}}^2 + {\sigma_{e^o}}^2},
    \label{eq:ccc}
\end{equation}
where $\rho_{e^*e^o}$ denotes the Pearson's coefficient of correlation between $e^*$ and $e^o$, and $(\mu_{e^*}, \mu_{e^o}), (\sigma_{e^*}, \sigma_{e^o})$ denote their mean and standard deviation values, respectively.

\noindent{\textbf{AU Intensity Estimation}} is evaluated using two standard metrics: Intra-class Correlation Coefficient (ICC) and Mean Square Error (MSE), computed for each AU individually.

\subsection{Training Details}
\noindent \textbf{Loss Functions.} To train the dimensional emotion recognition models, we use inverse-CCC + MSE loss, following the objective function originally proposed in ~\cite{Kossaifi_2020_CVPR}. Whereas for the AU intensity estimation, we use the MSE alone as the loss function, similar to the existing methods~\cite{sanchez2021affective,ntinou2021transfer}. In both cases, the per-frame loss is accumulated over an input image sequence in computing the total loss per mini-batch.

\noindent \textbf{Optimisation. } We use the Adam optimiser~\cite{kingma2014adam} to train all the models evaluated in this work. Note that in the 2D appearance baselines, unlike in the case of 3D face models, CNN backbones and GRU blocks are trained end-to-end. During training, dropout values of GRUs and FC layers are set to 0.5 and 0.25 respectively. Each mini-batch is composed of 4 sequences, with each sequence containing 100 frames. The initial learning rate value is 1e-4 and it is tuned using a cosine annealing based scheduler with warm restarts enabled~\cite{loshchilov2016sgdr}. Also, $L_2$ regularisation is applied during model training by setting the weight decay value to 1e-4.

\section{Results and Discussion} \label{sec:res}

\begin{table}[]
    \centering
    \renewcommand{\arraystretch}{1.3} 
    \begin{tabular}{l c c c}
    \toprule
    \textbf{Method} & \textbf{Mean} & \textbf{Std. Dev} & \textbf{Median} \\
    & \textbf{(in mm)} & \textbf{(in mm)} & \textbf{(in mm)} \\
    \midrule
     ExpNet (3DMM-CNN~\cite{tuan2017regressing}) & 2.33 & 2.05 & 1.84 \\
     \A{RingNet~\cite{RingNet:CVPR:2019}} & \A{1.98} & \A{1.77} & \A{1.50} \\
     3DDFA-V2~\cite{Guo:2020:3DDFA_v2} & 1.57 & 1.39 & 1.23 \\
     DECA~\cite{Feng:2021:DECA} & 1.38 & 1.18 & 1.09 \\
    EMOCA~\cite{Danecek:2022:EMOCA}$\dagger$ & 1.38 & 1.18 & 1.09 \\
     \bottomrule
    \end{tabular}
    \caption{Monocular 3D face reconstruction error (scan-to-mesh distance) values reported in the leaderboard of NoW~\cite{zielonka2022towards,RingNet:CVPR:2019} evaluation repository ($\dagger$EMOCA has the same reconstruction performance as DECA)}
    \label{tab:now}
\end{table}

\begin{figure*}[h!]
\begin{center}
\includegraphics[width=1.0\linewidth]{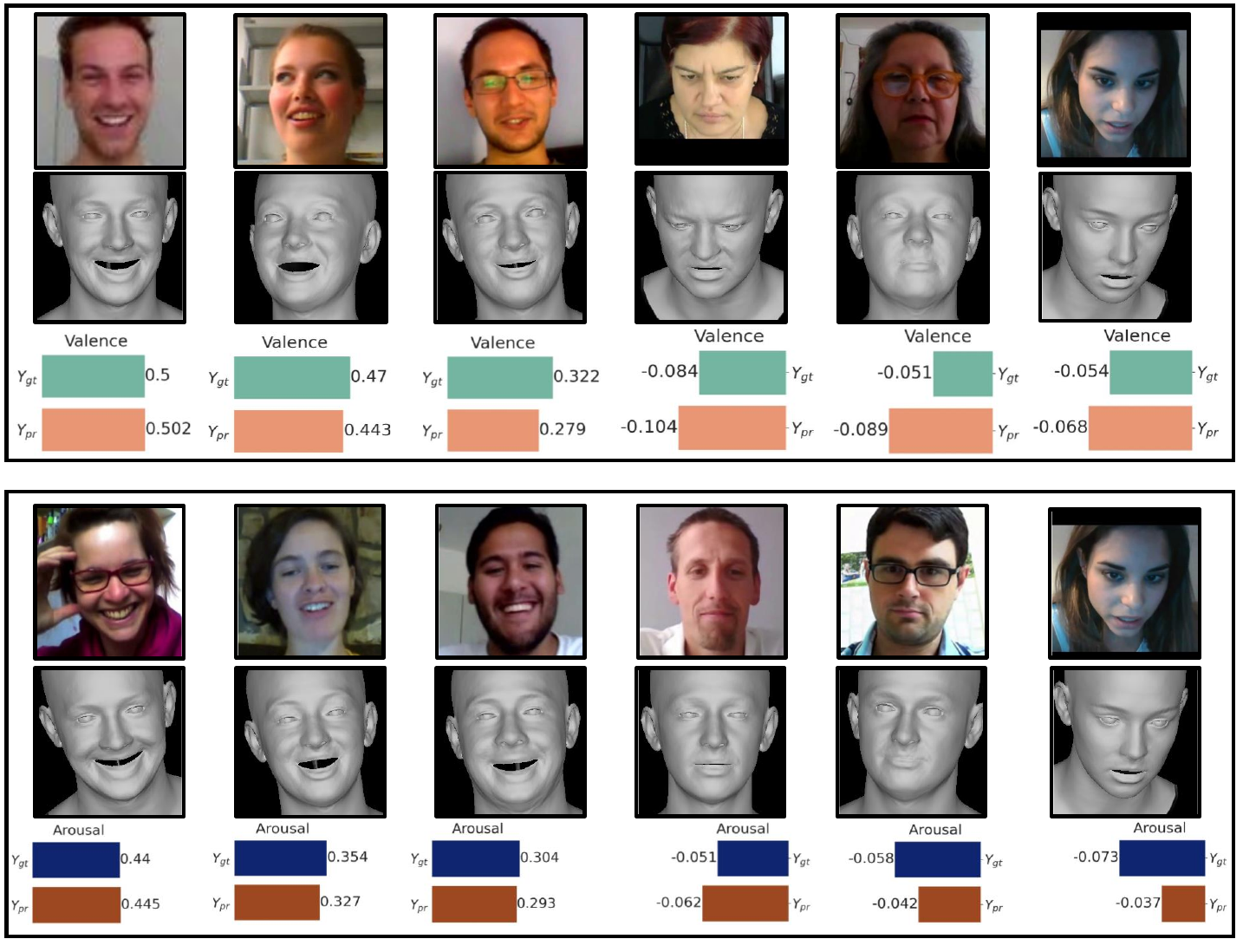}
\end{center}
\caption{Dimensional emotion recognition results on AVEC'19 validation examples using EMOCA~\cite{Danecek:2022:EMOCA} ($Y_{gt}$ and $Y_{pr}$ denote the ground truth and prediction values, respectively).}
\label{fig:qual_avec19}
\end{figure*}

\begin{figure*}[h!]
\begin{center}
\includegraphics[width=0.9\linewidth]{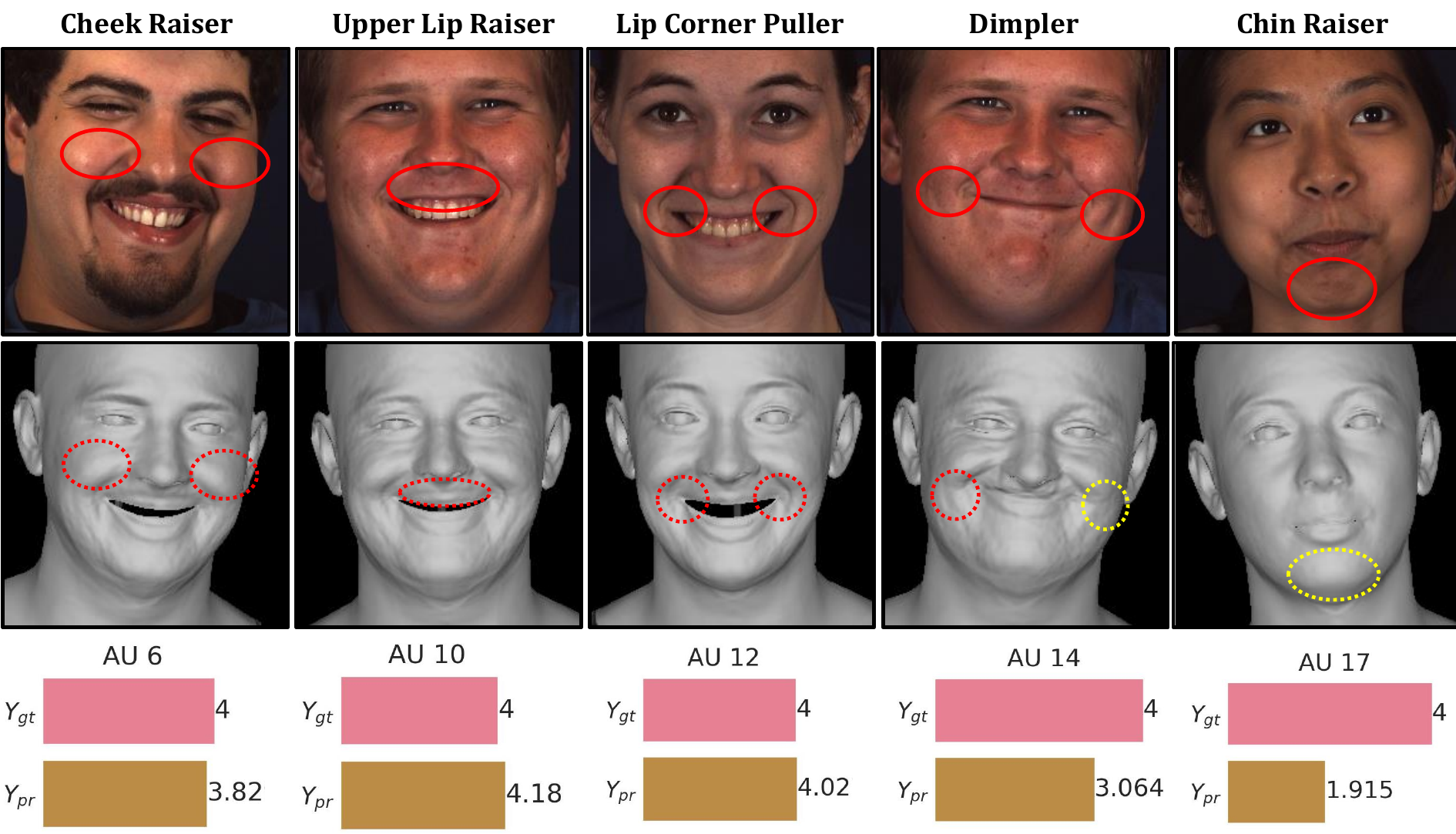}
\end{center}
   \caption{Action Unit intensity estimation results on BP4D validation examples using  EMOCA~\cite{Danecek:2022:EMOCA}  ($Y_{gt}$ and $Y_{pr}$ denote the ground truth and prediction values, respectively). Facial regions of AUs with low errors in the intensity prediction and with better 3D reconstruction quality are enclosed in red colored ellipses. The regions enclosed in yellow coloured ellipses highlight somewhat poorly reconstructed facial regions of AUs with high errors in the intensity prediction.}
\label{fig:qual_bp4d}
\end{figure*}

\subsection{Task-wise Performance Analysis} \label{subsection:taskwise-results}

\begin{table}
\begin{center}
\renewcommand{\arraystretch}{1.3} 
\begin{tabular}{l l c c c}
\toprule 
& & \textbf{Valence} & \textbf{Arousal} & \textbf{Avg.} \\
\textbf{Features} & \textbf{Model} & \textbf{CCC $\uparrow$} & \textbf{CCC $\uparrow$} & \textbf{CCC $\uparrow$} \\
\midrule
\multirow{6}{*}{\shortstack{\A{2D} \\ \A{Appearance}}} & Mitenkova et al.~\cite{mitenkova2019valence} 			  & 0.469 & 0.392 & 0.430 \\
& Toisoul et al.~\cite{toisoul2021estimation}  & 0.650 & 0.610 & 0.630 \\
& Kossaifi et al.~\cite{Kossaifi_2020_CVPR}	& 0.750 & 0.520 & 0.635\\
& APs~\cite{sanchez2021affective}	& 0.750 & 0.640  & 0.695  \\
\noalign{\vskip 1mm}
\cdashline{2-5}
\noalign{\vskip 1mm} 
& ResNet-50+GRU$^\dagger$  	 & 0.550 &  0.552 & 0.551 \\
& EmoFAN+GRU$^\dagger$ 	  & 0.715 & 0.568 & 0.641 \\
\cmidrule{2-5}
\multirow{5}{*}{\A{3D Shape}}  & ExpNet+GRU & 0.638 & 0.510 & 0.574 \\
& \A{RingNet+GRU} & \A{0.587} & \A{0.442} & \A{0.514}\\
& 3DDFA-V2+GRU & 0.710 & 0.646 & 0.678 \\
& DECA+GRU & 0.755 & 0.682 & 0.718 \\
& EMOCA+GRU & \textbf{0.775} & \textbf{0.716} & \textbf{0.745} \\
\bottomrule
\end{tabular}
\end{center}
\caption{Dimensional emotion recognition results on the SEWA test set ($^\dagger$ denotes in-house evaluation).}
\label{tab:sewa_results}
\end{table}

\begin{table}
    \centering
    \renewcommand{\arraystretch}{1.3}    
    \begin{tabular}{l l c c c}
    \toprule
    & & \textbf{Valence} & \textbf{Arousal} & \textbf{Avg.} \\
    \textbf{Features} & \textbf{Model} & \textbf{CCC $\uparrow$} & \textbf{CCC $\uparrow$} & \textbf{CCC $\uparrow$} \\
    \midrule
    \multirow{3}{*}{\shortstack{\A{2D} \\ \A{Appearance}}} & Zhao et al.$^*$~\cite{zhao2019adversarial} & 0.579 & \textbf{0.594} & \textbf{0.586}  \\
    \noalign{\vskip 1mm}
    \cdashline{2-5}
    \noalign{\vskip 1mm}     
    & ResNet-50+GRU$^\dagger$ & 0.495 & 0.522 & 0.508 \\
    & EmoFAN+GRU$^\dagger$ & 0.527 & 0.564 & 0.545 \\ 
    \cmidrule{2-5}
    \multirow{5}{*}{\A{3D Shape}}  & ExpNet+GRU & 0.534 & 0.505 & 0.519 \\
    & \A{RingNet+GRU} & \A{0.540} & \A{0.557} & \A{0.548} \\
    & 3DDFA-V2+GRU & \textbf{0.590} & 0.544 & 0.567 \\
    & DECA+GRU & 0.561 & 0.565 & 0.563\\
    & EMOCA+GRU & 0.580 & 0.568 & 0.574 \\
    \bottomrule 
    \end{tabular}
    \caption{Dimensional emotion recognition results on the AVEC'19 validation set. ($^*$denotes visual-only results reported in the AVEC'19 CES Winners~\cite{zhao2019adversarial} and $^\dagger$ denotes in-house evaluation)}
    \label{tab:avec19_results}
\end{table}

\begin{table*}
    \centering
    \renewcommand{\arraystretch}{1.3} 
    \begin{tabular}{l l l c c c c c c c}
    \toprule
    \textbf{Metric} & \textbf{Features} & \textbf{Model} & \textbf{6} & \textbf{10} & \textbf{12} & \textbf{14} & \textbf{17}  & \textbf{Avg.} \\
    \midrule
    \multirow{12}{*}{ICC $\uparrow$} & \multirow{6}{*}{\shortstack{\A{2D} \\ \A{Appearance}}}
    & CDL~\cite{baltruvsaitis2015cross}             & 0.69 & 0.73 & 0.83 & 0.50 & 0.37 & 0.62 \\ 
    & & ISIR~\cite{nicolle2015facial}                       & 0.79 & 0.80 & \textbf{0.86} & \textbf{0.71}  & 0.44 & 0.72 \\
    & & HR~\cite{ntinou2021transfer}                                & \textbf{0.82} & \textbf{0.82} & 0.80 & \textbf{0.71} & 0.50 & 0.73 \\
    & & APs~\cite{sanchez2021affective}													& \textbf{0.82} & \textbf{0.80} & \textbf{0.86} & 0.69 & \textbf{0.51} &\textbf{0.74}\\
    \noalign{\vskip 1mm}
    \cdashline{3-9}
    \noalign{\vskip 1mm}     
    & & ResNet-18+GRU & 0.75 & 0.71 & 0.79 & 0.63 & 0.45 & 0.66\\
    & & EmoFAN+GRU$^\dagger$ & 0.78 & 0.76 & 0.83 & 0.62 & 0.50 &  0.70 \\
    \cmidrule{3-9}
    & \multirow{6}{*}{\A{3D Shape}}  & ExpNet+GRU & 0.57 & 0.56 & 0.69 & 0.35 & 0.38 & 0.51 \\
    & & \A{RingNet+GRU} & \A{0.70} & \A{0.70} &    \A{ 0.79} & \A{0.29} & \A{0.28} & \A{0.57} \\
    & & 3DDFA-V2+GRU & 0.73 & 0.67 & 0.87 & 0.36 & 0.31 & 0.59 \\
    & & DECA+GRU &  0.72 & 0.68 & 0.83 & 0.42 & 0.23 & 0.58\\
    & & EMOCA+GRU & 0.73 & 0.68 & 0.86 & 0.34 & 0.27 & 0.58 \\
     \midrule
      \multirow{12}{*}{MSE $\downarrow$} & \multirow{6}{*}{\shortstack{\A{2D} \\ \A{Appearance}}}
       & CDL~\cite{baltruvsaitis2015cross}                  &  -   &  -   &  -   &  -   &  -   &  -   \\ 
       & &  ISIR~\cite{nicolle2015facial}                      		& 0.83 & 0.80 & 0.62 & 1.14 & 0.84 & 0.85  \\
       & & HR~\cite{ntinou2021transfer}                                		 & \textbf{0.68}  & \textbf{0.80} & 0.79 & \textbf{0.98} & 0.64 & 0.78\\     
        & & APs~\cite{sanchez2021affective}                                          				& 0.72  & 0.84 &\textbf{0.60} & 1.13 &\textbf{ 0.57} &  \textbf{0.77}\\
\noalign{\vskip 1mm}
\cdashline{3-9}
\noalign{\vskip 1mm}         
       & & ResNet-18+GRU$^\dagger$ & 0.81 & 0.90 & 0.82 & 1.17 & 0.82 & 0.91 \\
       & & EmoFAN+GRU$^\dagger$ & 0.79 & 0.85 & 0.76 & 1.19 & 0.78 & 0.87 \\
    \cmidrule{3-9}
    & \multirow{6}{*}{\A{3D Shape}}   & ExpNet+GRU & 1.5 & 1.56 & 1.33 & 1.59 & 0.88 & 1.37 \\
    & & \A{RingNet+GRU} &    \A{0.90} &    \A{1.05} & \A{0.91} & \A{1.59} & \A{0.90} & \A{1.07} \\
    & & 3DDFA-V2+GRU & 0.91 & 1.22 & 0.61 & 1.48 & 0.86 & 1.01 \\
    & & DECA+GRU & 0.98 & 1.14 & 0.8 & 1.57 & 1.18 & 1.13 \\
    & & EMOCA+GRU & 0.82 & 1.08 & 0.6 & 1.75 & 0.96 & 1.04 \\
    \bottomrule 
    \end{tabular}
    \caption{BP4D test set results ($^\dagger$denotes in-house evaluation).}
    \label{tab:bp4d_test_results}
\end{table*}

\noindent \textbf {Dimensional Emotion Recognition on SEWA.} Table~\ref{tab:sewa_results} presents the results of valence and arousal estimation on the SEWA test set in three groups: recent state-of-the-art (SOTA) benchmarks, in-house evaluated 2D CNN baselines using ResNet-50 and EmoFAN backbones, and 3D face models. Note that GRU modules of the same modelling capacity are used on top of the in-house evaluated 2D CNN features as well as 3D face features. We can clearly see that EMOCA features outperform all the remaining models listed in Table~\ref{tab:sewa_results} by considerable margins in terms of both valence and arousal dimensions. Let us consider the best performing 2D CNN baseline, EmoFAN, as a reference with the CCC values of 0.715 and 0.568 for valence and arousal respectively. EMOCA expression features outperform EmoFAN by +.060 and +.148 in valence and arousal CCC scores respectively. EMOCA even outperforms all the existing benchmarks on SEWA with improved mean CCC values in the range of +.315~\cite{mitenkova2019valence} to +.050~\cite{sanchez2021affective}. Interestingly,  RingNet has a lower mean CCC score compared to ExpNet, although its 3D reconstruction error is higher than that of ExpNet (see Table~\ref{tab:now}). DECA and 3DDFA-V2 improve mean CCC values by +.077 and +.037 compared to the mean CCC of EmoFAN. Further, we observe that 3DDFA-V2 is on par with the SOTA method (APs~\cite{sanchez2021affective}), which benefits from a more advanced temporal learning method based on stochastic context modelling.  

To summarise, the above-discussed results on SEWA show that the 3D face features are expressive enough to recognise continuous emotions and they perform better than 2D appearance features used in existing state-of-the-art models. Note that in contrast to all 2D appearance-based approaches that applied transfer learning to CNN backbones, the training procedure of 3D face models does not rely on any labelled facial expression data. One exception is EMOCA, which used AffectNet~\cite{mollahosseini2017affectnet} pretraining and an additional emotion recognition module in its training. Nevertheless, the remaining 3D face models that do not use any emotion labels in learning their expression parameters, perform on par or better than 2D appearance baselines.  \\

\noindent \textbf {Dimensional Emotion Recognition on AVEC'19 CES.}  Results on the AVEC'19 CES database, as shown in Table~\ref{tab:avec19_results}, exhibit similar trends. All five 3D face models perform far above the ResNet-50+GRU baseline, except for ExpNet; the rest also outperform a stronger 2D CNN baseline, EmoFAN+GRU. The AVEC'19 CES challenge winners, Zhao et al.~\cite{zhao2019adversarial} achieve a mean CCC score of +.012 above the best performing 3D face model, i.e. EMOCA. In valence estimation, unlike in the case of SEWA, the best performing method is 3DDFA-V2. Also, CCC values of ExpNet and RingNet models are in line with their 3D reconstruction errors shown in Table~\ref{tab:now}.

\Cref{fig:qual_avec19} illustrates the qualitative results of EMOCA in valence-arousal estimation on some of the validation examples from the AVEC'19 corpus. Interestingly, the emotion recognition performance is slightly worse in negative valence and arousal cases compared to their positive counterparts. This could be due to the availability of fewer training examples for negative valence and low arousal quadrants~\cite{kossaifi2019sewa, ringeval2019avec}.

\setlength{\tabcolsep}{5pt} 
\renewcommand{\arraystretch}{0.9} 
\begin{table*}
    \centering
    \renewcommand{\arraystretch}{1.3} 
    \begin{tabular}{c c l c c c c c c c c c c c c c}
    \toprule
 \textbf{Metric} & \textbf{Features} & \textbf{Model} &   1       &      2    &    4      &     5    &    6     &    9   &   12  &  15  &    17  &  20   &   25  &  26  & Avg.\\
    \midrule
     \multirow{16}{*}{ICC $\uparrow$} &  \multirow{6}{*}{\shortstack{\A{2D} \\ \A{Appearance}}}
     & G2RL~\cite{fang2rl} 												& \textbf{0.71}  & 0.31     & \textbf{0.82} & 0.06    & 0.48 & \textbf{0.67} & 0.68 & 0.21 & 0.47 & 0.17 &\textbf{0.95} &  \textbf{0.75} & 0.52 \\
     & & RE-Net~\cite{yang2020re}									   & 0.59    & \textbf{0.63}   & 0.73 & \textbf{0.82}    & 0.49 & 0.50 & 0.73 & 0.29 & 0.21 & 0.03  & 0.90 & 0.60 & 0.54 \\     
    & & VGP-AE~\cite{eleftheriadis2016variational}  &  0.48    &  0.47     &  0.62  & 0.19   & 0.50   & 0.42 & 0.80 & 0.19 & 0.36 & 0.15 & 0.84 & 0.53 & 0.46 \\
    & & 2DC~\cite{linh2017deepcoder}   				  	 & 0.70     &  0.55    &  0.69  &  0.05   &\textbf{0.59}  & 0.57 &\textbf{0.88}& 0.32 & 0.10 & 0.08 & 0.90 & 0.50 & 0.50\\
    & & HR~\cite{ntinou2021transfer}    								     &  0.56    &  0.52    &  0.75  &  0.42   &  0.51  & 0.55 & 0.82 &\textbf{0.55}& 0.37 & 0.21 &0.93& 0.62 & 0.57 \\
    & & APs~\cite{sanchez2021affective}    														    & 0.35     &  0.19   & 0.78   &    0.73  & 0.52 & 0.65& 0.81   & 0.49 &\textbf{0.61}&\textbf{0.28}& 0.92 & 0.67 & \textbf{0.58}              		\\    
\noalign{\vskip 1mm}
\cdashline{3-16}
\noalign{\vskip 1mm}     
    & & ResNet-18+GRU$^\dagger$ & 0.21& 0.16& 0.71& 0.65& 0.55& 0.59& 0.78& 0.41& 0.54& 0.22& 0.90& 0.64 & 0.53\\             		 
    & & EmoFAN+GRU$^\dagger$                         & 0.23     &  0.13     &  0.77 &  0.70    & 0.53  & 0.64 &  0.82 & 0.42 & 0.58 & 0.25 & 0.92 &0.69&  0.56              			\\
    \cmidrule{3-16}
    &  \multirow{6}{*}{\shortstack{\A{3D} \\ \A{Shape}}} & ExpNet+GRU & -0.03 & -0.07 & 0.16 & -0.02 & 0.25 & 0.12 & 0.38 & 0.04 & 0.09 & -0.01 & 0.57 & 0.31 & 0.15 \\
    & & \A{RingNet+GRU} & \A{0.0} & \A{0.01} & \A{0.09} & \A{0.03} & \A{0.45} & \A{0.21} & \A{0.66} & \A{0.07} & \A{0.0} & \A{-0.01} & \A{0.88} & \A{0.43} & \A{0.23} \\ 
    & & 3DDFA-V2+GRU & 0.17 & 0.23 & 0.19 & 0.0 & 0.52 & 0.32 & 0.78 & 0.01 & -0.01 & 0.02 & 0.78 & 0.42 & 0.28 \\
    & & DECA+GRU & 0.09 & -0.03 & 0.33 & 0.07 & 0.54 & 0.30 & 0.79 & 0.14 & 0.14 & 0.0 & 0.73 & 0.33 & 0.29 \\
    & & EMOCA+GRU& 0.31 & 0.17 & 0.76 & 0.57 & 0.48 & 0.52 & 0.85 & 0.21 & 0.16 &  -0.01 & 0.84 & 0.28 & 0.43 \\
     \midrule
     \multirow{16}{*}{MSE $\downarrow$} & \multirow{8}{*}{\shortstack{\A{2D} \\ \A{Appearance}}}
      & G2RL~\cite{fang2rl} &  -  & - & - & - & - & - & - & - & - & - & - & - & -\\
     & & RE-Net~\cite{yang2020re} &  -  & - & - & - & - & - & - & - & - & - & - & - & - \\           
     & & VGP-AE~\cite{eleftheriadis2016variational}      & 0.51 & \textbf{0.32} & 1.13 & 0.08 & 0.56 & 0.31 & 0.47 & 0.20 & 0.28 & \textbf{0.16} & 0.49        & 0.44        & 0.41 \\
     & & 2DC~\cite{linh2017deepcoder}                  		   & \textbf{0.32} & 0.39 & 0.53 & 0.26 & 0.43 & 0.30 & \textbf{0.25} & 0.27 & 0.61 & 0.18 & 0.37      & 0.55       & 0.37 \\   
     & & HR~\cite{ntinou2021transfer}	                         		           & 0.41 & 0.37 & 0.70 & 0.08 & 0.44 & 0.30 & 0.29 & 0.14  & 0.26 & \textbf{0.16} & 0.24          & 0.39          & 0.32 \\ 
     & & APs~\cite{sanchez2021affective}                                          						  &  0.68 & 0.59 &\textbf{0.40} & \textbf{0.03} & 0.49 & \textbf{0.15} & 0.26 & \textbf{0.13} & 0.22 & 0.20 & 0.35 & \textbf{0.17} & \textbf{0.30} \\
\noalign{\vskip 1mm}
\cdashline{3-16}
\noalign{\vskip 1mm} 
     & & ResNet-18+GRU$^\dagger$ & 0.88& 0.71& 0.54& 0.13& 0.38& 0.26& 0.39& 0.20& 0.28& 0.25& 0.32& 0.41 & 0.39 \\
     & & EmoFAN+GRU$^\dagger$         				 &  0.85 & 0.79 & 0.48 & 0.06 & 0.47 & 0.19 & 0.34 & 0.18 & 0.23 & 0.21 & 0.30          & 0.40          & 0.37 \\ 
    \cmidrule{3-16}
    & \multirow{6}{*}{\shortstack{\A{3D} \\ \A{Shape}}} & ExpNet+GRU & 0.93 & 0.99 & 1.77 & 0.09 & 0.74 & 0.57 & 1.09 & 0.20 & 0.37 & 0.18 & 1.43 & 0.6 & 0.75\\
    & & \A{RingNet+GRU} & \A{0.55} & \A{0.47} & \A{1.32} & \A{0.07} & \A{0.45} & \A{0.30} & \A{0.53} & \A{0.15} & \A{0.33} & \A{0.16} & \A{0.39} & \A{0.43} & \A{0.44}\\ 
    & & 3DDFA-V2+GRU & 0.53 & 0.39 & 1.39 & 0.07 & 0.57 & 0.27 & 0.50 & 0.20 & 0.33 & 0.14 & 0.83 & 0.54 & 0.48 \\
    & & DECA+GRU & 0.61 & 0.65 & 2.23 & 0.08 & 0.45 & 0.40 & 0.39 & 0.18 & 0.37 & 0.17 & 0.92 & 0.50 & 0.57 \\
    & & EMOCA+GRU& 0.66 & 0.64 & 0.63 & 0.05 & 0.53 & 0.29 & 0.28 & 0.18 & 0.36 & 0.20 & 0.52 & 0.56 & 0.41 \\ 
    \bottomrule
    \end{tabular}
    \caption{Aggregated 3-fold cross validation results on DISFA dataset ($^\dagger$ denotes in-house evaluation).}
    \label{tab:disfa_results}
\end{table*}

\begin{table}
    \centering
    \renewcommand{\arraystretch}{1.3} 
    \begin{tabular}{l l l}
    \toprule
    \midrule
     \multirow{5}{*}{Group 1: 0.6 $<$ ICC $<$ 1.0} 
    & AU 4 & Brow Lowerer \\
    & AU 12 & Lip Corner Puller \\
    & AU 10 & Upper Lip Raiser \\
    & AU 25 & Lips Part \\
    & AU 6 &  Cheek Raiser \\
    \midrule
    \multirow{5}{*}{Group 2: 0.4 $<$ ICC $<$ 0.6} 
    & AU 5 & Upper Lid Raiser \\
    & AU 9 & Nose Wrinkler \\
    & AU 14 & Dimpler \\
    & AU 26 & Jaw Drop \\
    \midrule
    \multirow{5}{*}{Group 3: ICC $<$ 0.4} 
    & AU 1 & Inner Brow Raiser \\
    & AU 2 & Outer Brow Raiser \\
    & AU 15 & Lip Corner Depressor \\
    & AU 17 & Chin Raiser \\
    & AU 20 & Lip stretcher \\
    \midrule
    \bottomrule 
    \end{tabular}
    \caption{Segregation of AUs, from both BP4D and DISFA, according to their best performance with 3D face expression features.} 
    \label{tab:au_groups}
\end{table}

\noindent \textbf{AU Intensity Estimation on BP4D.} Table~\ref{tab:bp4d_test_results} presents the results of existing SOTA benchmarks, our in-house evaluated 2D CNNs and 3D face models, on the test set of BP4D. By comparing average ICC and MSE values achieved by the models listed in Table~\ref{tab:bp4d_test_results}, we can clearly notice that 3D face models have inferior performance to 2D appearance-based baselines. In contrast to the results of valence-arousal models, in estimating AU intensities all 3D face features, including EMOCA, fall behind 2D appearance features. 

Among the 3D face models, ExpNet achieves the poorest results in terms of average ICC and MSE scores. Particularly, all five 3D face models show consistently inferior performance in predicting the intensities of AU 14 and AU 17. \Cref{fig:qual_bp4d} qualitatively illustrates the performance of EMOCA on the BP4D test set examples. In this illustration, we can notice a clear correspondence between less accurate predictions made for AUs such as dimpler (AU 14) and chin raiser (AU 17) and somewhat poor 3D reconstructions of their corresponding facial regions (enclosed in yellow coloured ellipses in \Cref{fig:qual_bp4d}). For instance, in the case of AU 17, the details of the chin region are poorly reconstructed in its 3D face, which could explain the poor performance of EMOCA expression features in predicting AU 17 in this example. \\

\noindent \textbf{AU Intensity Estimation on DISFA.} In Table~\ref{tab:disfa_results}, we compare the aggregated results of 3-fold cross-validation of different SOTA methods, 2D CNN baselines, and 3D face models. Similar to BP4D results, here also 3D face models perform inferior to 2D appearance-based models. Among 3D face models, ExpNet has the worst performance, and EMOCA has the best performance. Unlike in the case of BP4D, EMOCA achieves significantly better performance (ICC score of +.14 w.r.t. DECA) than the remaining 3D face models. Compared to the best performing 2D appearance-based model, APs~\cite{sanchez2021affective}, EMOCA has lower performance by a margin of -.15 ICC score. 

Based on the combined results on BP4D and DISFA datasets, we segregate all the AUs into three groups based on their best ICC values in the case of 3D face models. We observe that only five AUs (12, 10, 25, 6) listed in group 1 are captured well in the 3D face expressions. It seems, the subtler the AUs (e.g. AU 17 -- chin raiser) are, the worse their ICC scores are with 3D face models. One possible explanation is that such subtle expression-specific 3D reconstruction errors are likely to get suppressed by the global reconstruction loss functions that are commonly used in training 3D face alignment models. Further, the size of the facial region corresponding to each AU varies widely. \A{For example, as shown in \Cref{fig:qual_bp4d}, facial regions corresponding to AU10 (upper lip raiser) are wider than the areas corresponding to AU17 (chin raiser).}

Overall, the trends in AU estimation results clearly show that the 3D faces are still far from capturing the fine-grained facial expressions that are critical to fully understanding expressive facial behaviour. While the emotion recognition performance of all 3D shape features is considerably better than 2D appearance features, it is interesting to note their inability to recognise a wide range of action units. Towards explaining this discrepancy, we performed a correspondence analysis between AUs and emotion labels. 

\subsection{Correspondence Analysis: Dimensional Emotions and Action Units}
We investigate the significance of different AUs to recognising dimensional emotions, for reconciling the above observations regarding 3D face models' results on emotion and AU estimation tasks. To this end, we perform a simple linear regression analysis -- in which AU intensities are used as input features to predict their corresponding emotion labels. By comparing the coefficient values of different AUs, we interpret the importance of each AU in predicting the target emotion labels. For this purpose, we use a recently released in-the-wild emotion recognition corpus, Aff-wild-2~\cite{kollias2020analysing}, in which video data is annotated with both valence-arousal values and their corresponding AU occurrences. Although the Aff-Wild-2 dataset seems to be a more suitable candidate for the 3D face expression evaluation, it is composed of highly challenging, in-the-wild videos. As depicted in \Cref{fig:affw2}, the top 3 best performing 3D face models show poor reconstruction results on the Aff-wild-2 face images, except for EMOCA, which shows slightly better performance w.r.t. capturing facial expressions. Rather than pushing the limits of 3D face models to perform well in such in-the-wild conditions, our focus here is to investigate the current status of existing 3D face alignment models where they could attain acceptable shape-fitting performance.

\begin{figure}
\includegraphics[width=1.02\linewidth]{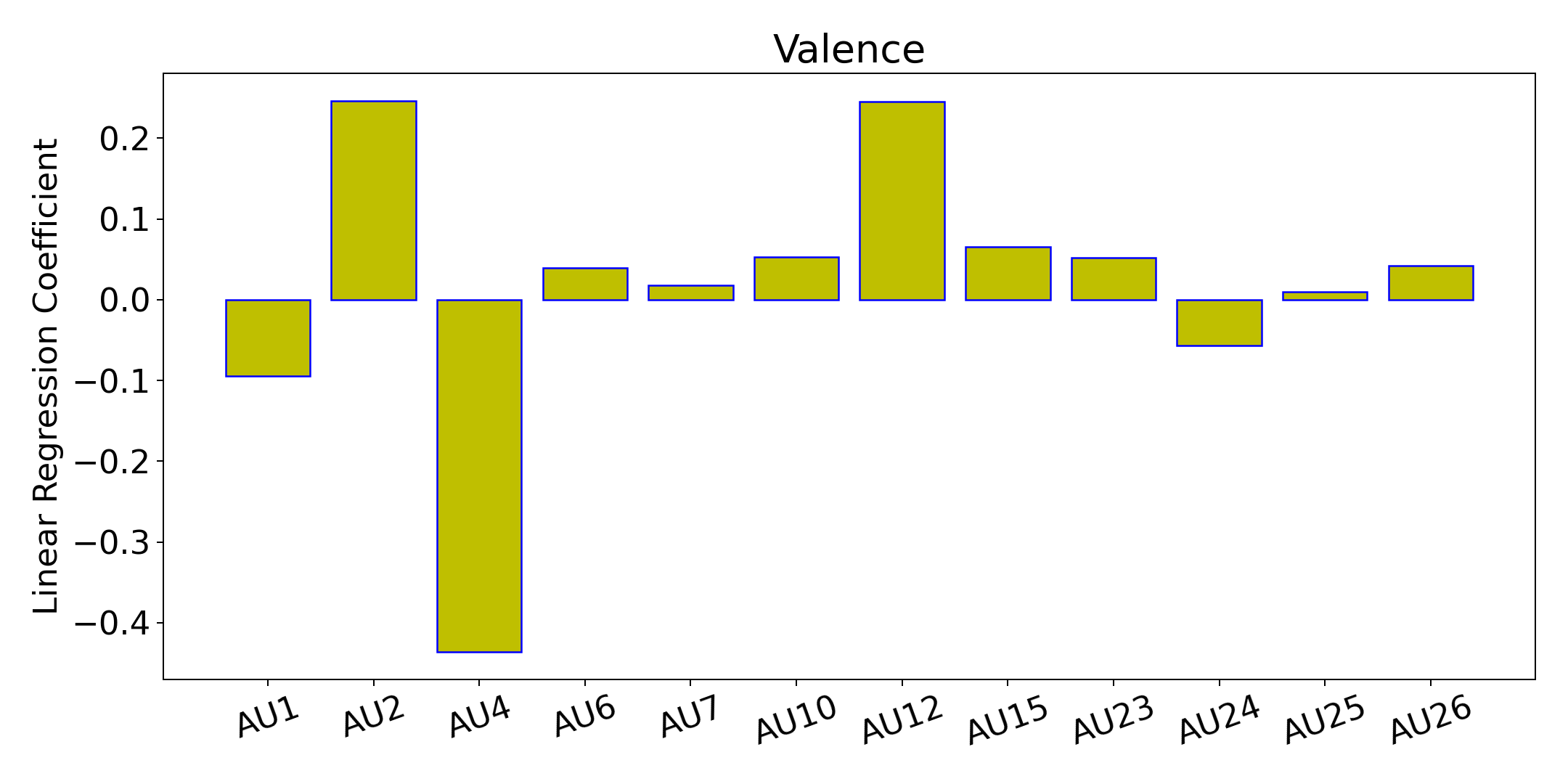}
   \caption{Coefficients of a linear regression model predicting \textbf{valence} from AUs.}
\label{fig:linear_vale}
\end{figure}

\begin{figure}
\includegraphics[width=1.02\linewidth]{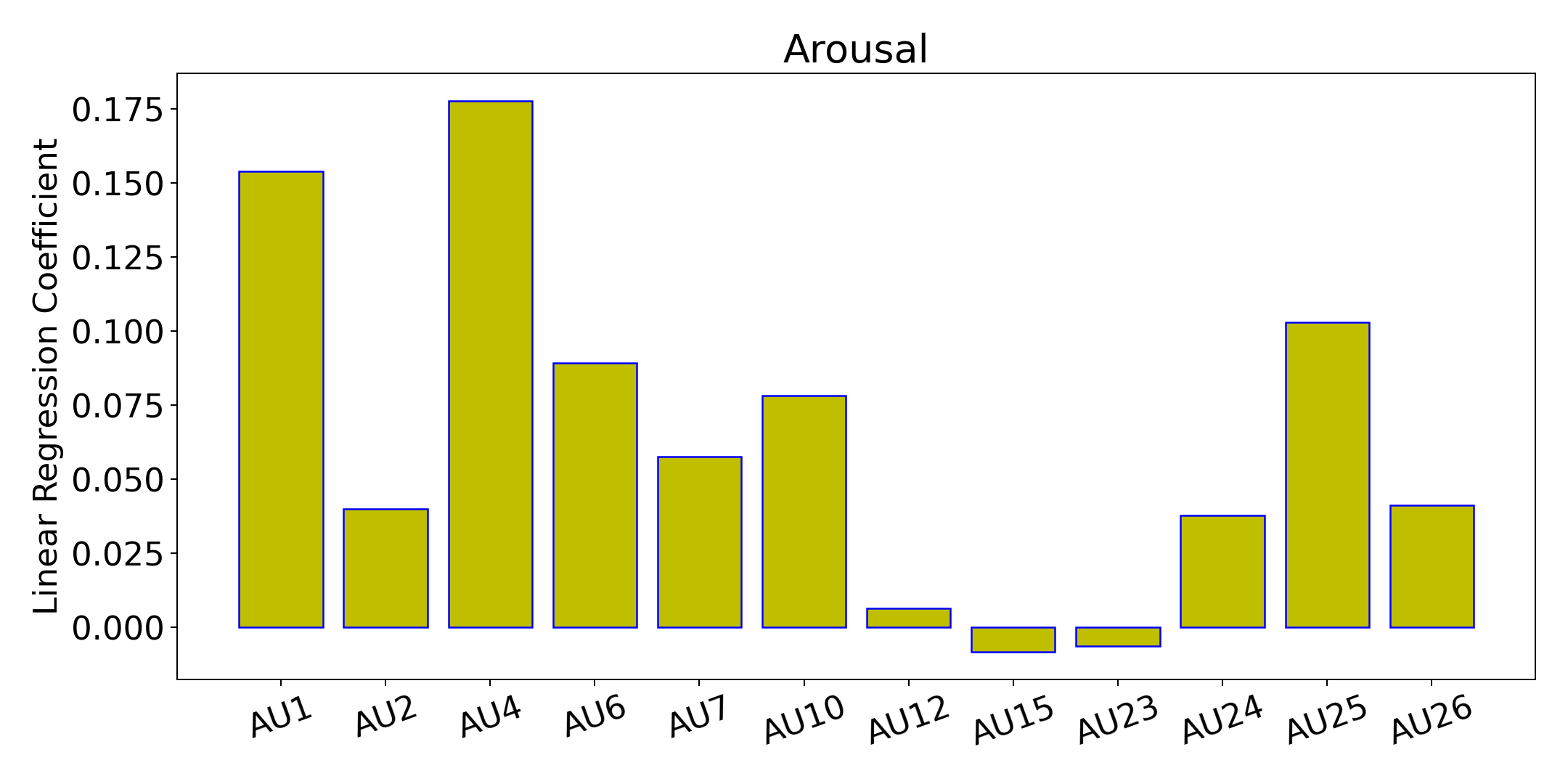}
   \caption{Coefficients of a linear regression model predicting \textbf{arousal} from AUs.}
\label{fig:linear_arou}
\end{figure}

\Cref{fig:linear_vale} and \Cref{fig:linear_arou} illustrate AU-wise regression coefficients for valence and arousal respectively. From these results, we can infer that the presence of AU 2 or AU 12 or the absence of AU 4 seems to be highly critical for predicting valence. Whereas for arousal prediction, the presence of AU 1 or AU 4 or AU 25 looks important. All five 3D face models perform well (see Group 1 in \Cref{tab:au_groups}) on at least one of the aforementioned AUs that seem critical for valence and arousal prediction. Thus, the superior emotion recognition performance of 3D face features is clearly explainable based on their relatively high ICC values for AU 4 (\Cref{tab:bp4d_test_results}) and AU 25 (\Cref{tab:disfa_results}).

\A{To further validate the efficacy of 3D face expression features in recognising apparent emotions, we  extend our experimental analysis to categorical emotion recognition. Below, we present a correspondence analysis between different AUs and discrete emotion classes. Refer to Appendix~\ref{appendix:disc_emo} for discrete emotion recognition results of 3D face features.}  

\begin{figure*}
    \centering
     \begin{subfigure}[]{0.49\textwidth}
         \centering
         \includegraphics[width=\textwidth]{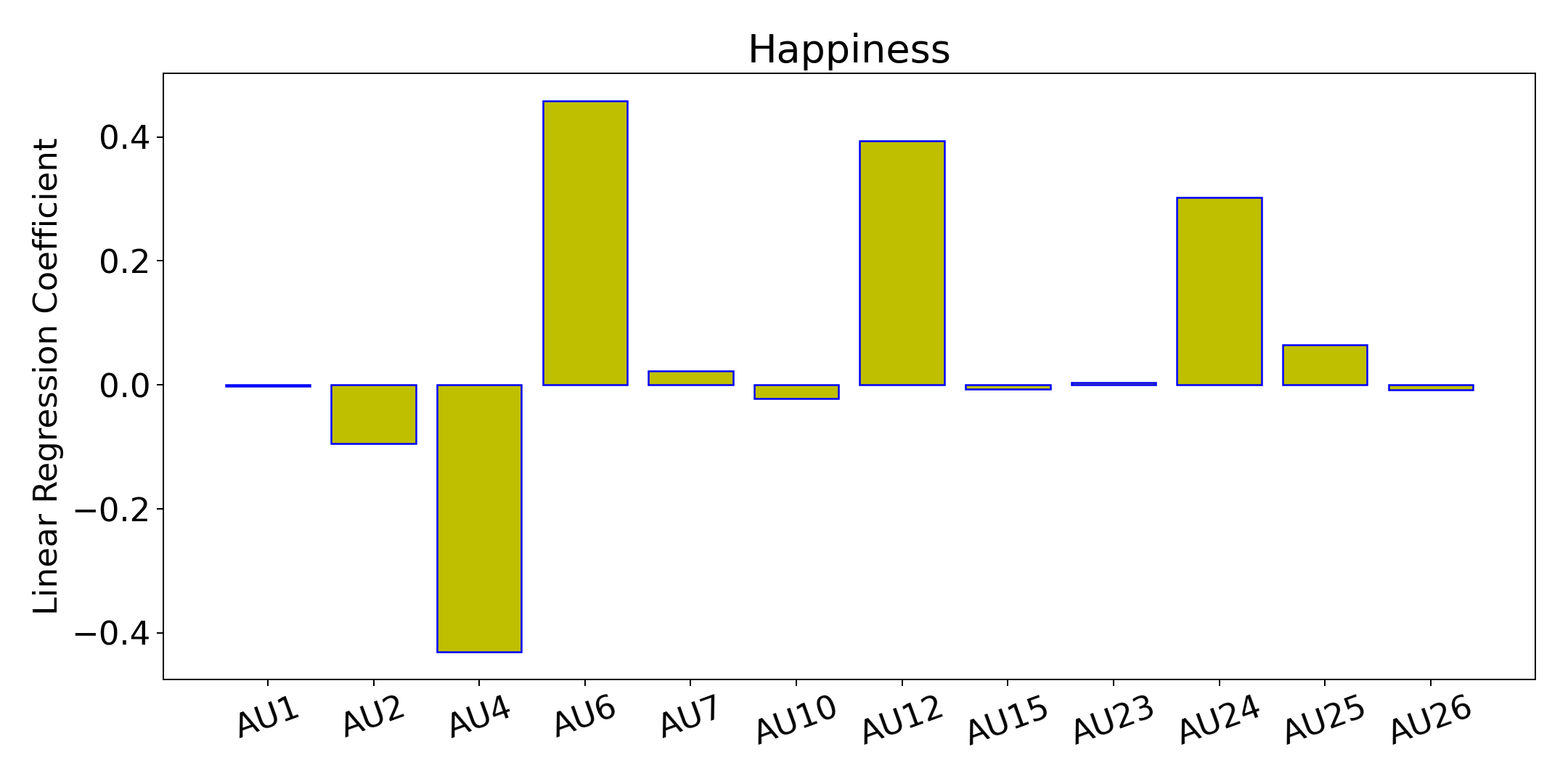}
     \end{subfigure}
     \begin{subfigure}[]{0.49\textwidth}
         \centering
         \includegraphics[width=\textwidth]{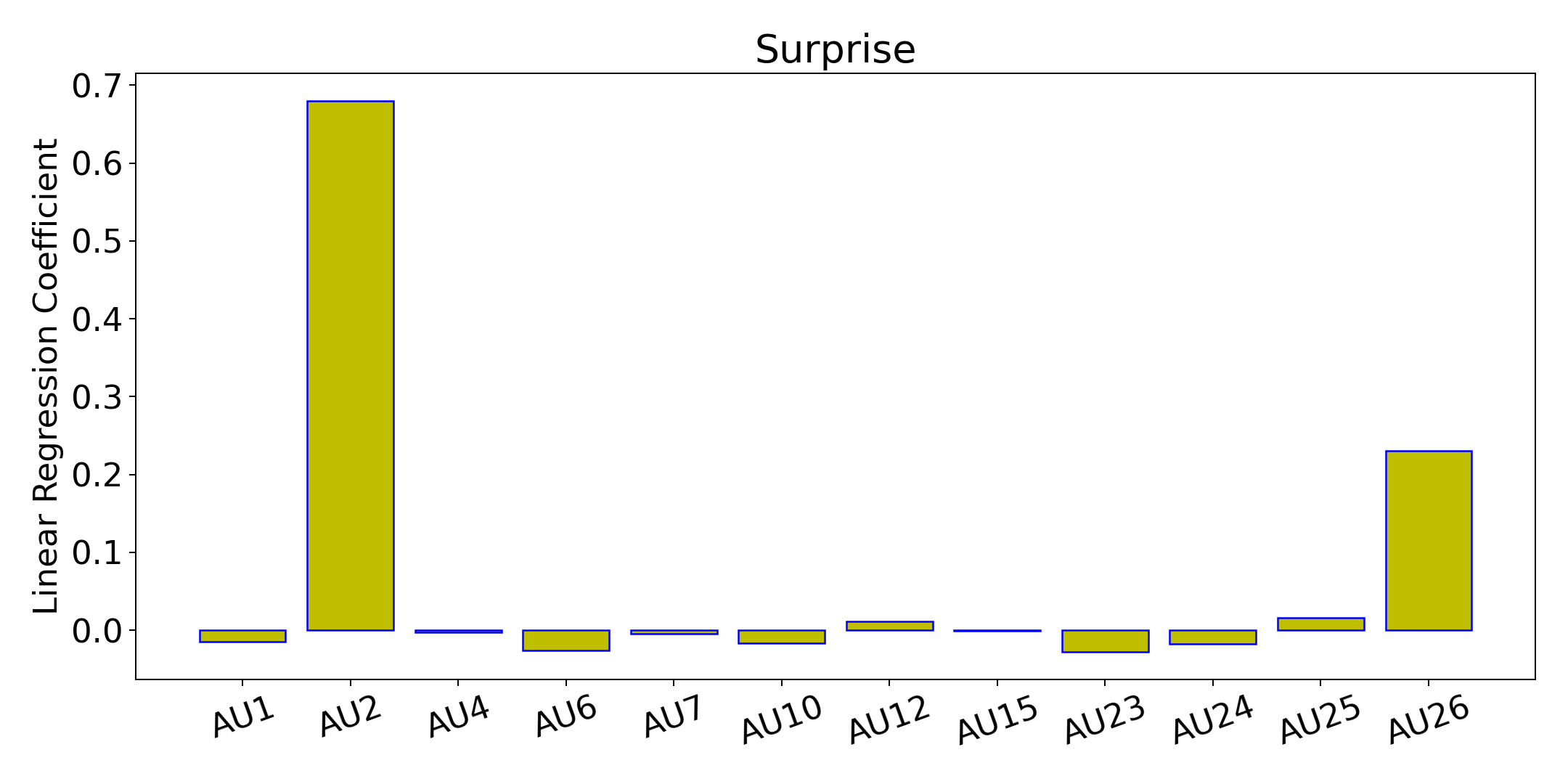}
     \end{subfigure}
     \begin{subfigure}[]{0.49\textwidth}
         \centering
         \includegraphics[width=\textwidth]{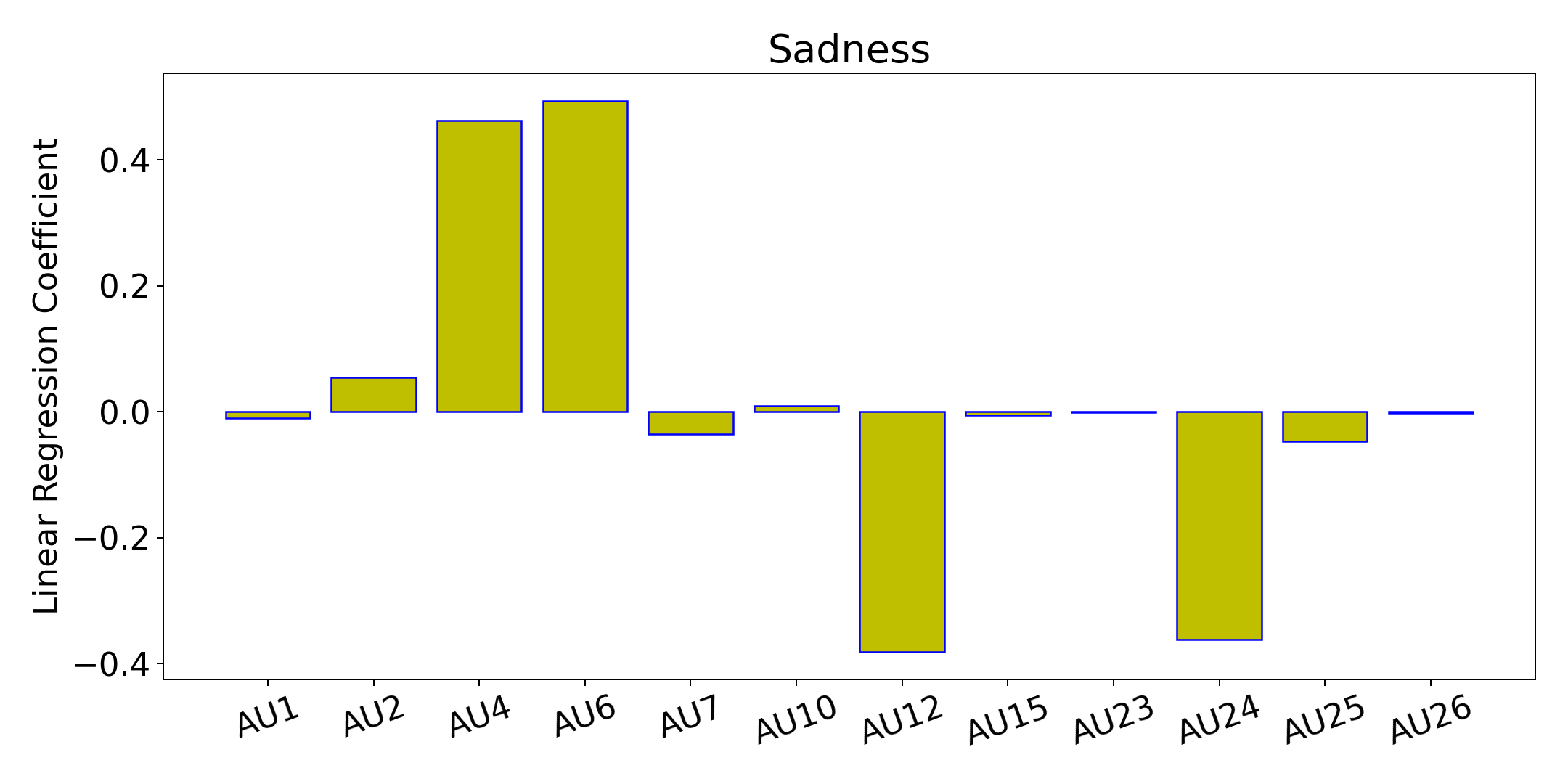}
     \end{subfigure}
     \begin{subfigure}[]{0.49\textwidth}
         \centering
         \includegraphics[width=\textwidth]{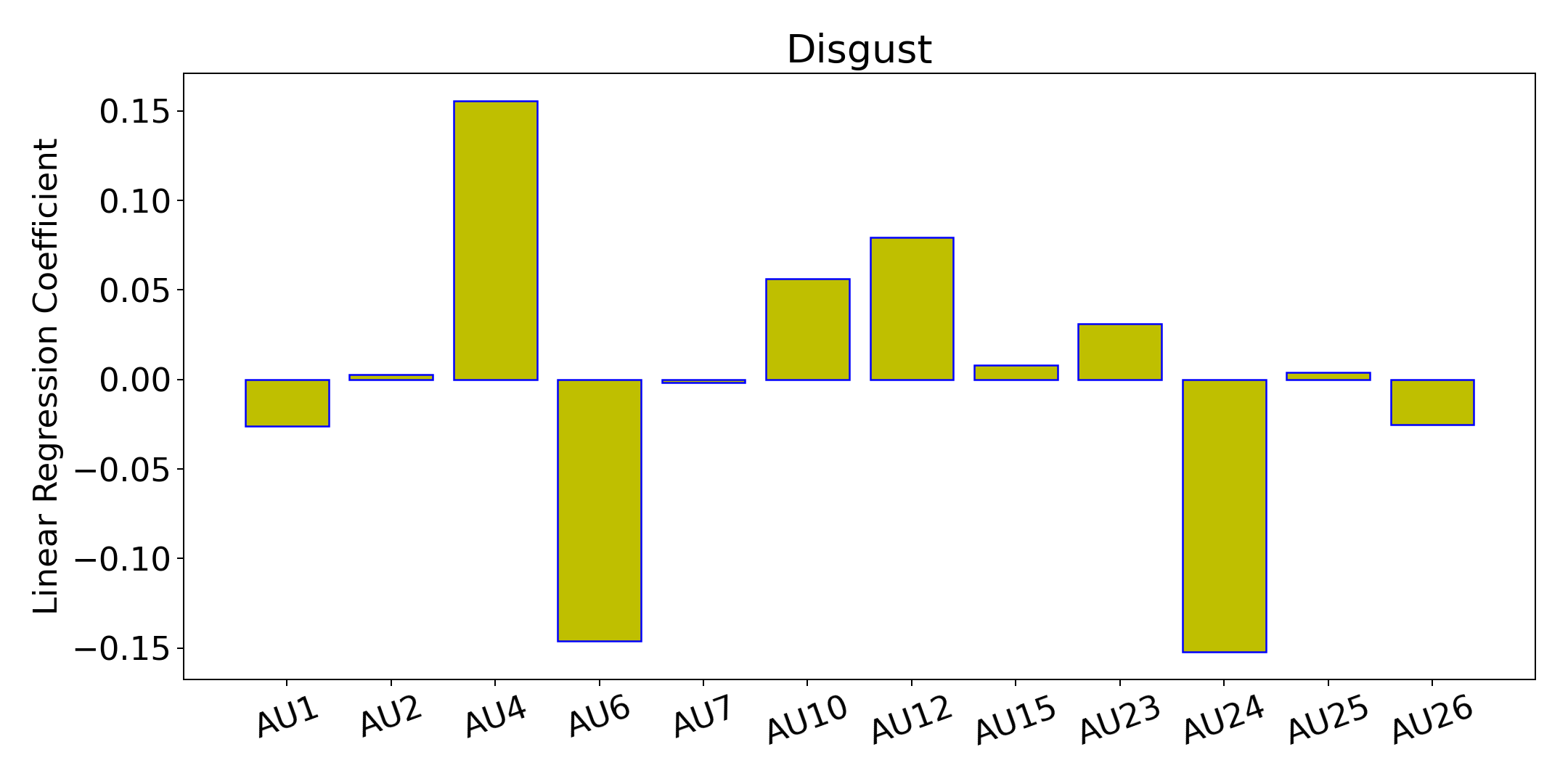}
     \end{subfigure}     
     \begin{subfigure}[]{0.49\textwidth}
         \centering
         \includegraphics[width=\textwidth]{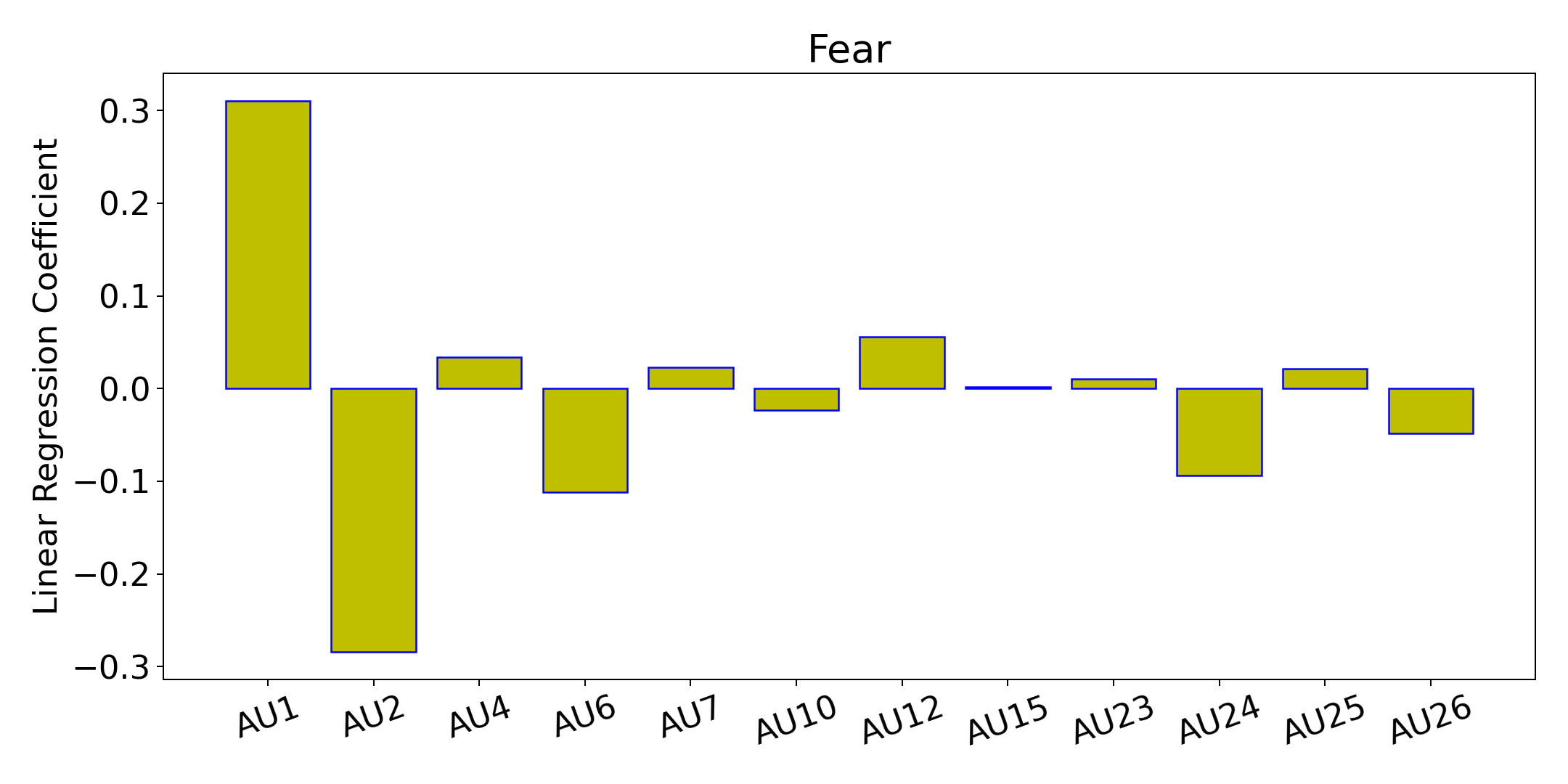}
     \end{subfigure}     
     \begin{subfigure}[]{0.49\textwidth}
         \centering
         \includegraphics[width=\textwidth]{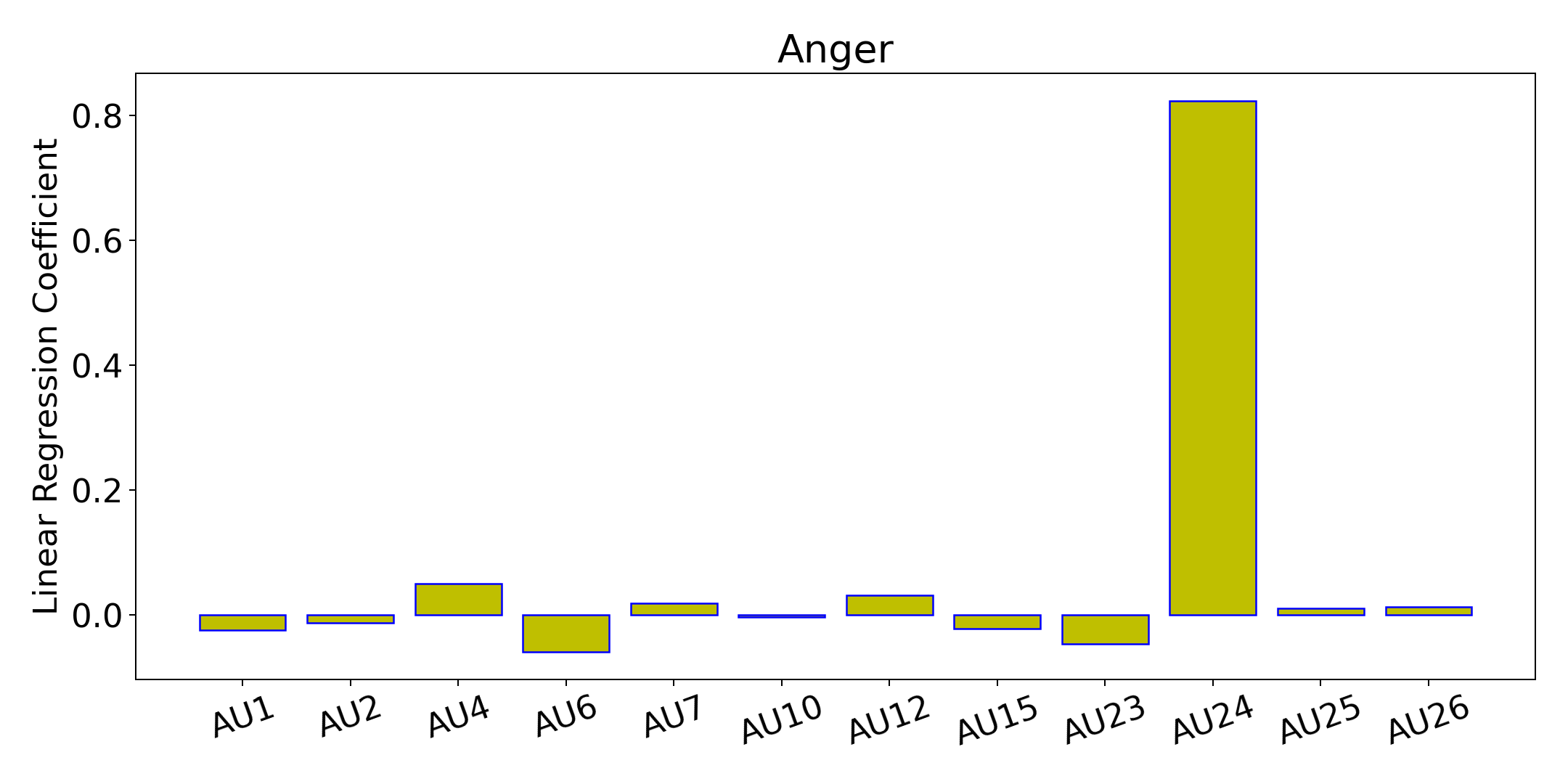}
     \end{subfigure}          
        \caption{Coefficients of linear regression model predicting \textbf{discrete emotions} from AUs.}
        \label{fig:linear_discEmo}
\end{figure*}

\A{\noindent \textbf{Correspondence Analysis between Discrete Emotions and AUs.} \Cref{fig:linear_discEmo} presents a comparison of emotion-wise regression coefficients of different AUs labelled in the Aff-wild-2 dataset. Similar to the case of continuous emotions, the performance of 3D face models on discrete emotion recognition can be clearly explained by their performance on some specific AUs. For instance, the emotion class 'happy' is strongly correlated with the presence of AU 6 or AU 12, or the absence of AU 4. For this class, 3D face models show the best recognition accuracy (see Figure~\ref{fig:disc-emo-cfee-results} and Figure~\ref{fig:disc-emo-ckplus-results} in Appendix~\ref{appendix:disc_emo}). Explaining their superior performance w.r.t.\ predicting the class 'happy', some of the 3D face models show good performance on AU 4 and AU 12 (see Group 1 in \Cref{tab:au_groups}). Similarly, the class 'fear' is correlated with the presence of AU 1 or AU 2, for which the 3D face models have very poor performance (see Group 3 in \Cref{tab:au_groups}).}

\subsection{Summary and Discussion} Based on all the above-discussed results on dimensional emotion recognition and AU intensity estimation tasks, and the correspondence analysis between apparent emotions and AU intensities, we draw the following conclusions:
\begin{itemize}
    \item 3D face shapes are expressive enough to regress dimensional representations of facial emotion. They are also good at capturing categorical emotion information (see Appendix.~\ref{appendix:disc_emo})
    \item But, in AU intensity estimation 3D face features are far from describing the complete set of facial actions and they fall behind the 2D appearance features.
\end{itemize}

\A{Except for the results of ExpNet and RingNet on SEWA}, the overall performance of 3D face features on emotion and AU intensity estimation is in line with their corresponding 3D shape reconstruction errors (see Table~\ref{tab:now}), as reported in the NoW evaluation repository leaderboard\footnote{Based on the challenge results provided at https://now.is.tue.mpg.de/nonmetricalevaluation.html. As EMOCA builds on the identity and shape encoders originally learned in DECA, they have the same reconstruction errors on the NoW evaluation repository.}. ExpNet, RingNet, 3DDFA-V2, and DECA do not use any emotion labels, and they are not trained on transferred representations from affect-related tasks. Deviating from these models, EMOCA, the best performing 3D face model, builds on DECA and it uses a valence-arousal estimation model pretrained on AffectNet~\cite{mollahosseini2017affectnet}. EMOCA additionally optimises an additional loss component, perceptual emotion consistency loss between the emotion features of RGB input and another valence-arousal estimator on the DECA model's expression and detail coefficients. It is important to note that our results show even through the use of AffectNet pretraining and emotion consistency, EMOCA demonstrates poor performance in capturing the facial actions corresponding to several AUs, as listed in Table~\ref{tab:au_groups}.


\A{For the poor AU intensity estimation performance of the 3D face models evaluated in this work, using a global basis vector for expression modelling could also be a reason. Alternative 3DMM formulations based on sparse and localised shape models, such as the ones proposed in ~\cite{neumann2013sparse} and ~\cite{Ferrari:2021}, could mitigate this problem to some extent. However, when applied to in-the-wild face image data, the generalisation performance of such sparse and locally constrained 3D face models has yet to be demonstrated in the literature. On the other hand, the benefits of current deep learning-based approaches used in this work are that they incorporated different auxiliary terms (i.e., face recognition, consistency losses, texture modelling, etc) in their optimisation and used large-scale in-the-wild 2D datasets for training.}

To make 3D face shape models expressive enough to capture the complete set of facial actions, discrete or continuous emotion labels alone as additional supervision signals do not suffice. It is important to focus on collecting 3D face scans captured in conditions eliciting individual AUs and their combinations. Towards addressing this challenge, it is also important to leverage naturally available supervision cues, such as temporal coherency of facial actions in a video~\cite{tellamekala2019temporally,lu2020self}, to learn more expressive 3D face shapes in a label-efficient manner.

Another important consideration is to increase the expressiveness of 3D face models by enhancing the representation capacity of 3DMMs. To this end, increasing the dimensionality of expression coefficient vector is one possible solution. However, as mentioned in several prior 3D face alignment works (e.g.~\cite{Guo:2020:3DDFA_v2}), higher dimensional expression vectors may negatively impact the shape reconstruction loss optimisation, hence slowing down the convergence of model training.  

\begin{figure}
\begin{center}
\includegraphics[width=1.0\linewidth]{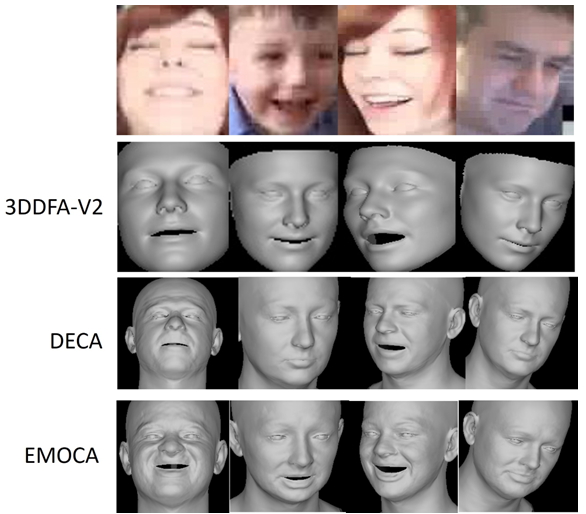}
\end{center}
   \caption{Shape fitting results of the top 3 best performing 3D face models for representative images sampled from Aff-wild 2 dataset.}
\label{fig:affw2}
\end{figure}

\noindent \textbf{Ethical Considerations and Limitations.} Automated facial expression analysis, particularly in affective computing, has valuable use cases for society. For example,  human-computer interaction, learning analytics, mental health and well-being and teleconferencing are only a few of these beneficial applications for facial expression analysis. However, there exist potential use cases raising ethical questions such as surveillance and military applications.

From the algorithmic fairness point of view, building emotion recognition based on 3D face models has advantages over CNN models that learn emotions directly from RGB images and videos. Most datasets are imbalanced in gender, ethnicity, age, and other appearance-relevant traits. Even though algorithmic bias is still a significant and open issue, learning from emotion coefficients of 3D face models discards all additional information that appearance-based CNN models jointly learn and condition on. However, 3D face models require good quality images for alignment (for instance, see the qualitative performance of all compared 3D face models in Fig.~\ref{fig:affw2}), which may limit their use cases.

\section{Conclusion}
\label{sec:conc}

We systematically investigated the ability of 3D face models to capture expression-induced shape deformations. By evaluating the 3D face expressions on the standard emotion recognition and AU intensity corpora, we presented a detailed exposition of their current strengths and limitations compared to state-of-the-art models based on 2D face image sequences. Our key findings in this study pointed out that expression features from 3D face models can achieve state-of-the-art results on time-continuous dimensional emotion recognition by outperforming most previous works and strong 2D face appearance baselines. However, the poor performance of 3D face models in AU intensity estimation indicates that their expression features are far from describing the complete set of facial actions.

\ifCLASSOPTIONcompsoc
  \section*{Acknowledgments}
\else
  \section*{Acknowledgment}
\fi

The work of Mani Kumar Tellamekala was supported  by  the  Engineering and  Physical Science  Research  Council project  (2159382)  and  Unilever  U.K. Ltd, and the work of Michel Valstar was supported by  the Nottingham Biomedical Research Centre (BRC). This work was also partially funded by the European Union Horizon 2020 research and innovation programme, grant agreement 856879 (Present), and the German Research Foundation DFG, grant agreement AN 559/8-1 (Panorama).

\ifCLASSOPTIONcaptionsoff
  \newpage
\fi

\bibliographystyle{IEEEtran}
\bibliography{egbib}

\clearpage

\begin{appendices}
\section{Discrete Emotion Recognition}
\label{appendix:disc_emo}
In discrete emotion recognition tasks, we evaluate the top 3 best performing 3D face alignment models considered in this work: 3DDFA-v2, DECA, and EMOCA, on the CK+~\cite{Lucey:2010} and CFEE~\cite{du2014compound} datasets. Both these datasets are acquired in controlled lab settings. Note that here our objective is not to aim for a state-of-the-art performance but to compare the expression representations derived from the 3D face models, as an ablation study.

\begin{figure}[t]
    \centering
     \begin{subfigure}[]{0.49\textwidth}
         \centering
         \includegraphics[width=\textwidth]{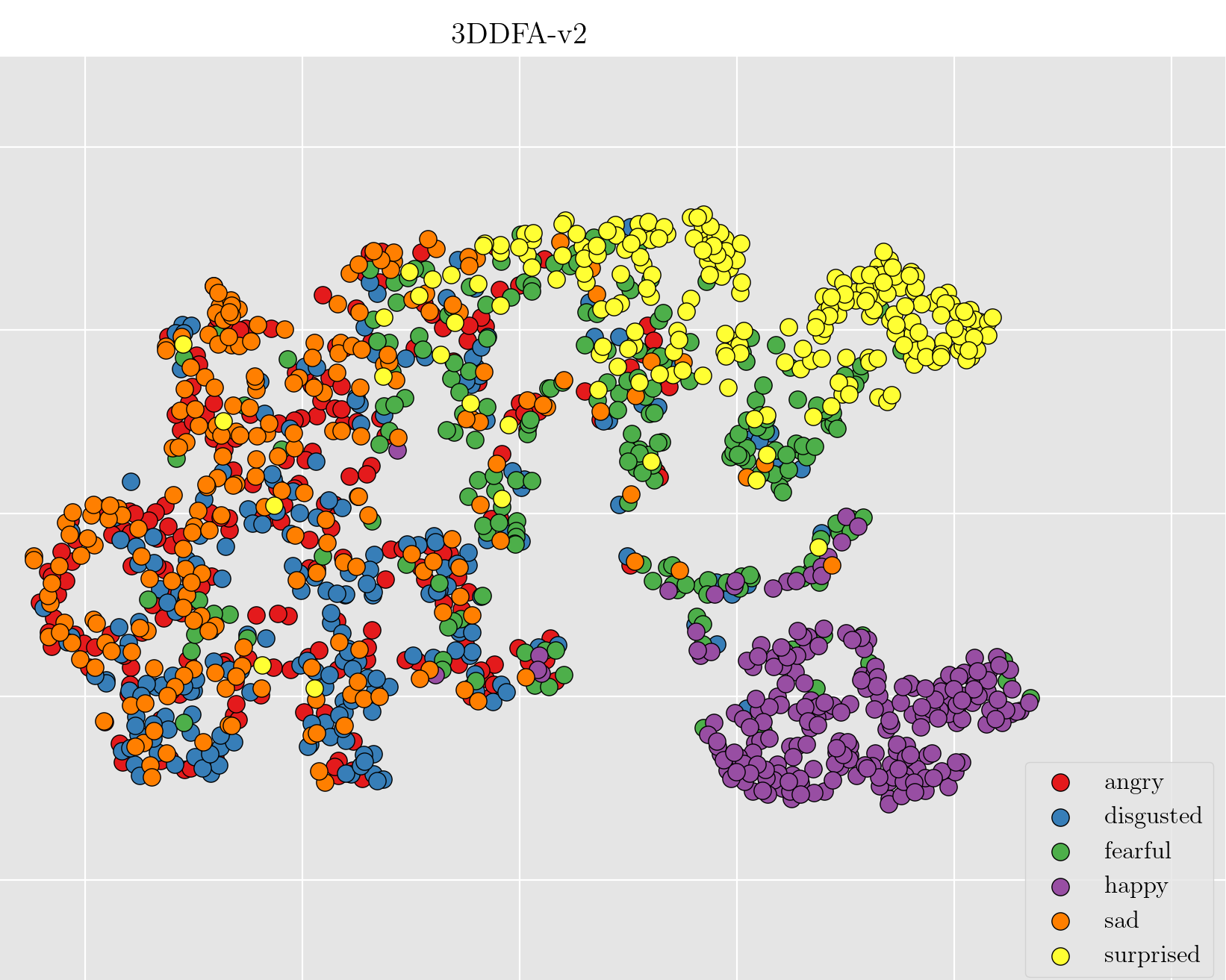}
     \end{subfigure}
     \begin{subfigure}[]{0.49\textwidth}
         \centering
         \includegraphics[width=\textwidth]{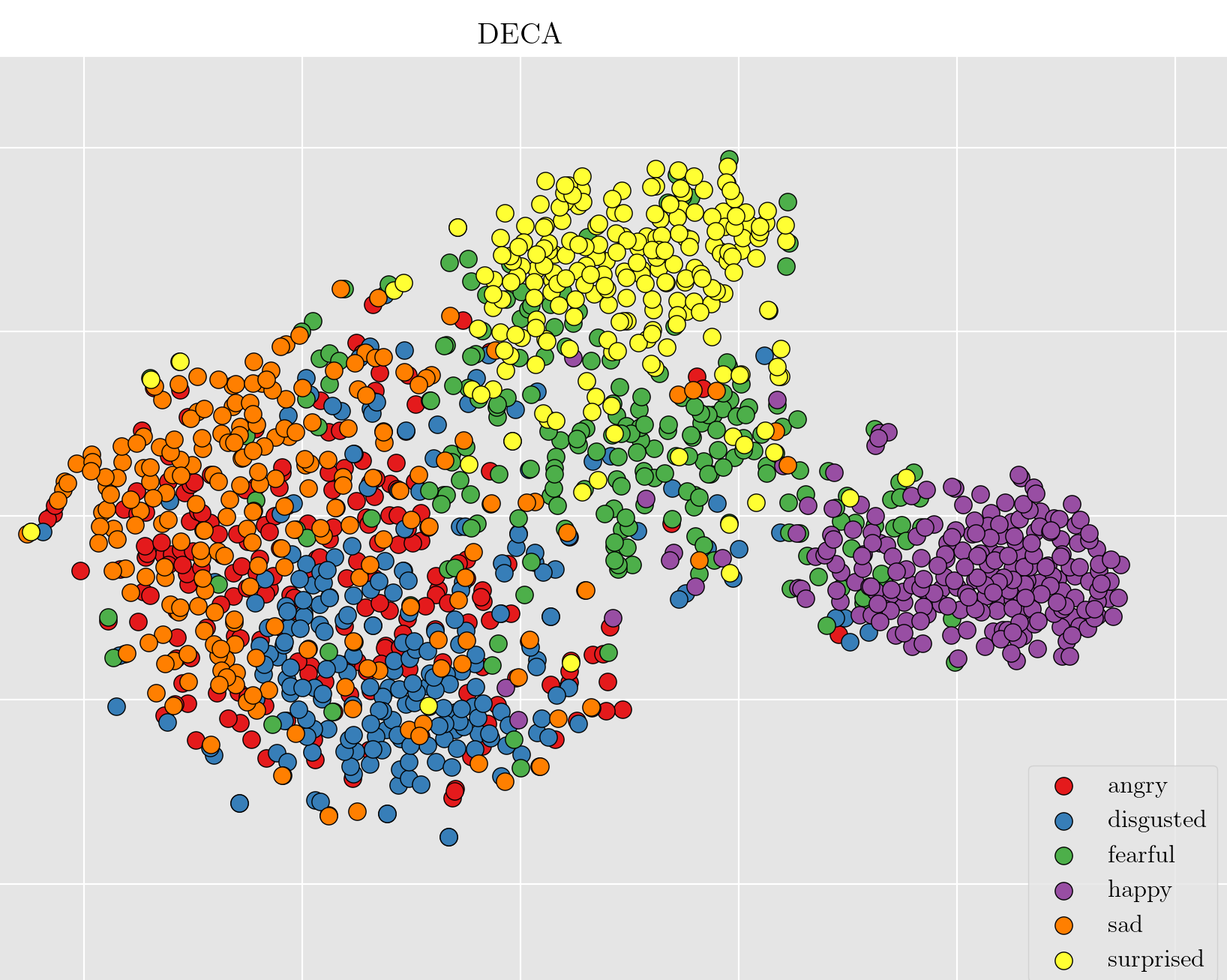}
     \end{subfigure}
     \begin{subfigure}[]{0.495\textwidth}
         \centering
         \includegraphics[width=\textwidth]{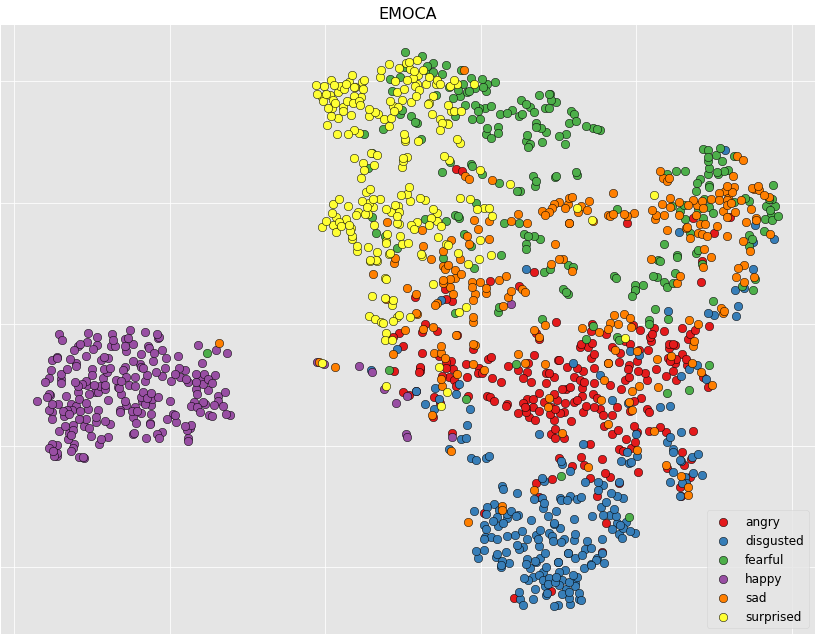}
     \end{subfigure}
     
    \caption{t-SNE distributions of the samples with basic emotions in CFEE database using 3DDFA-v2, DECA and EMOCA features}
    \label{fig:tsne}
\end{figure}

\begin{figure}
    \centering
     \begin{subfigure}[]{0.41\textwidth}
         \centering
         \includegraphics[width=\textwidth]{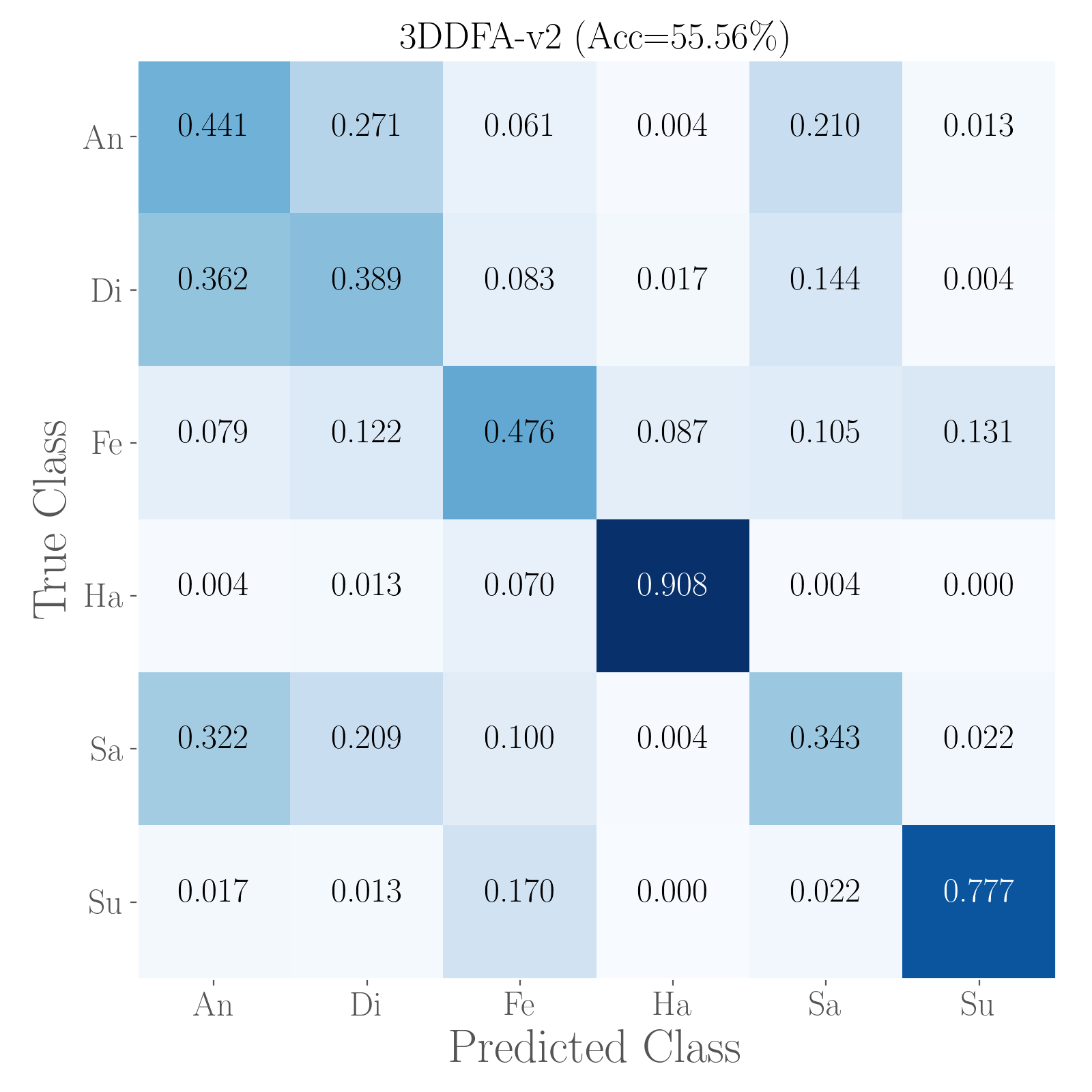}
     \end{subfigure}
     \begin{subfigure}[]{0.41\textwidth}
         \centering
         \includegraphics[width=\textwidth]{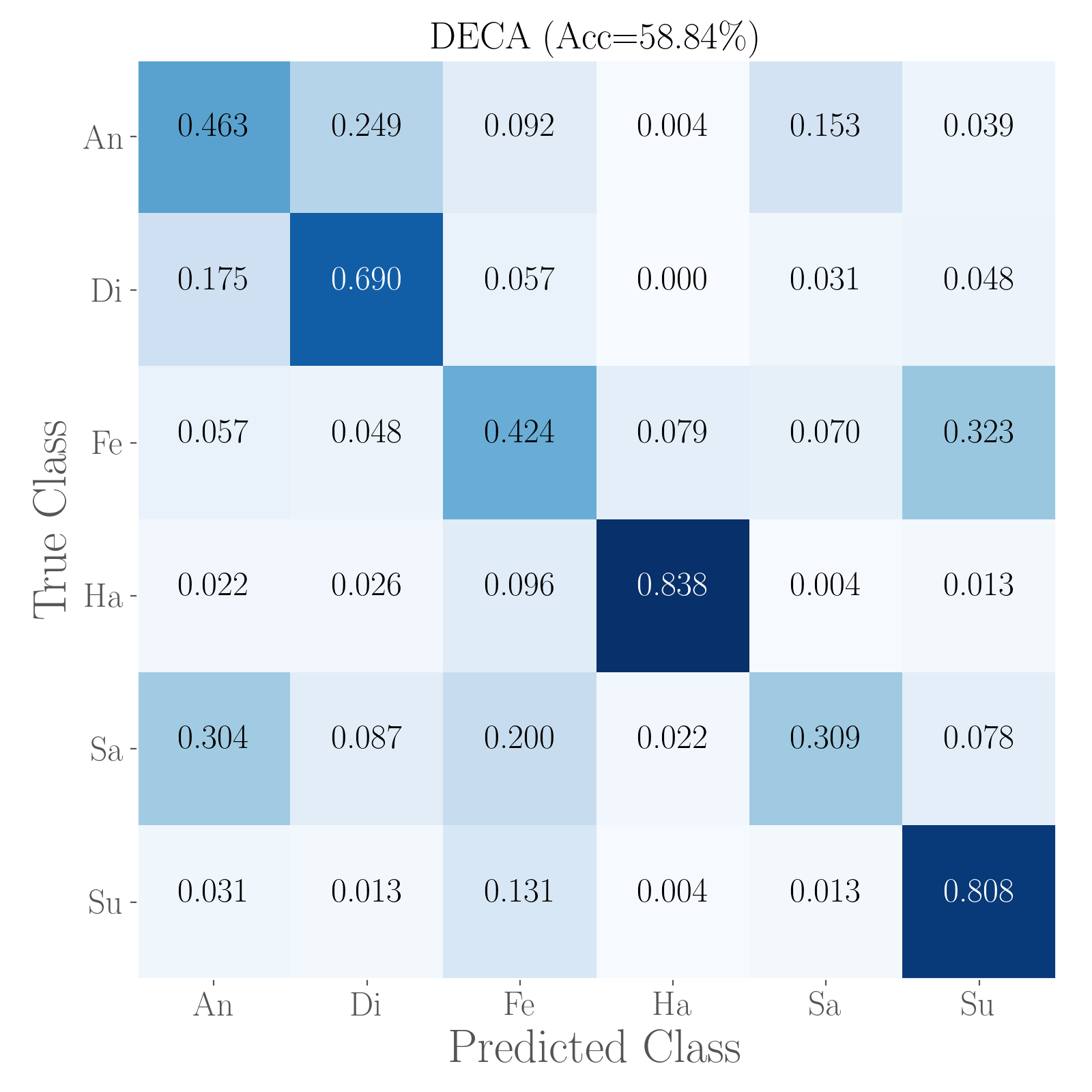}
     \end{subfigure}
     \begin{subfigure}[]{0.41\textwidth}
         \centering
         \includegraphics[width=\textwidth]{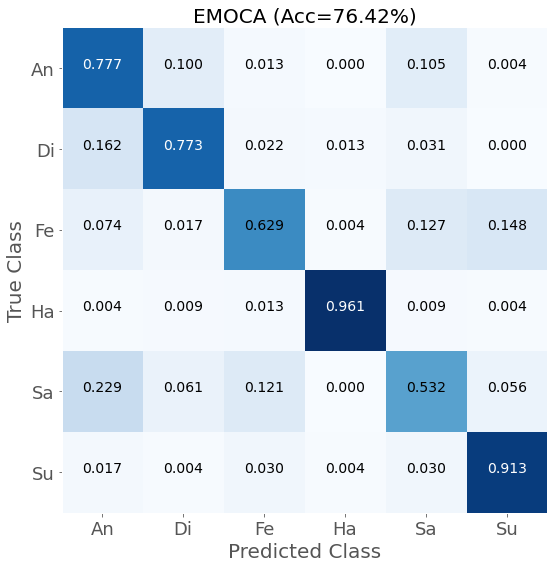}
     \end{subfigure}
     
    \caption{Discrete Emotion Recognition Results on \textbf{CFEE Dataset}}
    \label{fig:disc-emo-cfee-results}
\end{figure}

\textbf{CK+}~\cite{Lucey:2010} contains 327 video clips starting from a neutral state and ending at the apex point of anger (An), contempt (Co), disgust (Di), fear (Fe), happy (Ha), sadness (Sa), and surprise (Su). We use the apex frames in our evaluation. 

\textbf{CFEE}~\cite{Du:2014} contains still images of 230 subjects from diverse ethnic backgrounds with 22 basic and compound emotions categories. We us all samples (1375 images) labelled with basic emotions: anger (An), disgust (Di), fear (Fe), happy (Ha), sadness (Sa), and  surprise (Su).

As an ablation study to compare the expression coefficients of 3D face models, we normalise the expression coefficients according to the quantile range of the values and used a simple kNN classifier (k=5) with leave-one-out cross-validation and report the confusion matrices and emotion recognition accuracies.

\begin{figure}
    \centering
     \begin{subfigure}[]{0.41\textwidth}
         \centering
         \includegraphics[width=\textwidth]{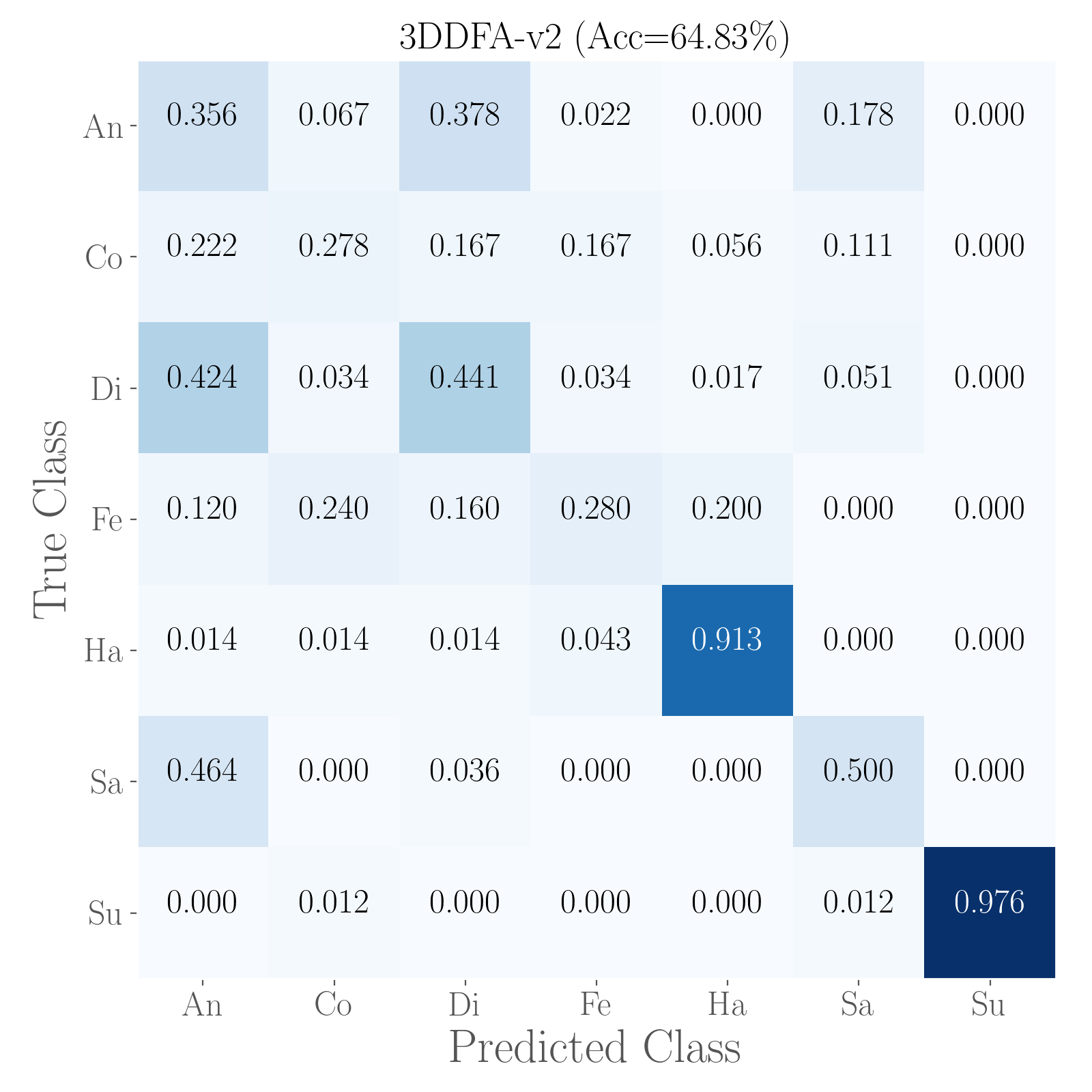}
     \end{subfigure}
     \begin{subfigure}[]{0.41\textwidth}
         \centering
         \includegraphics[width=\textwidth]{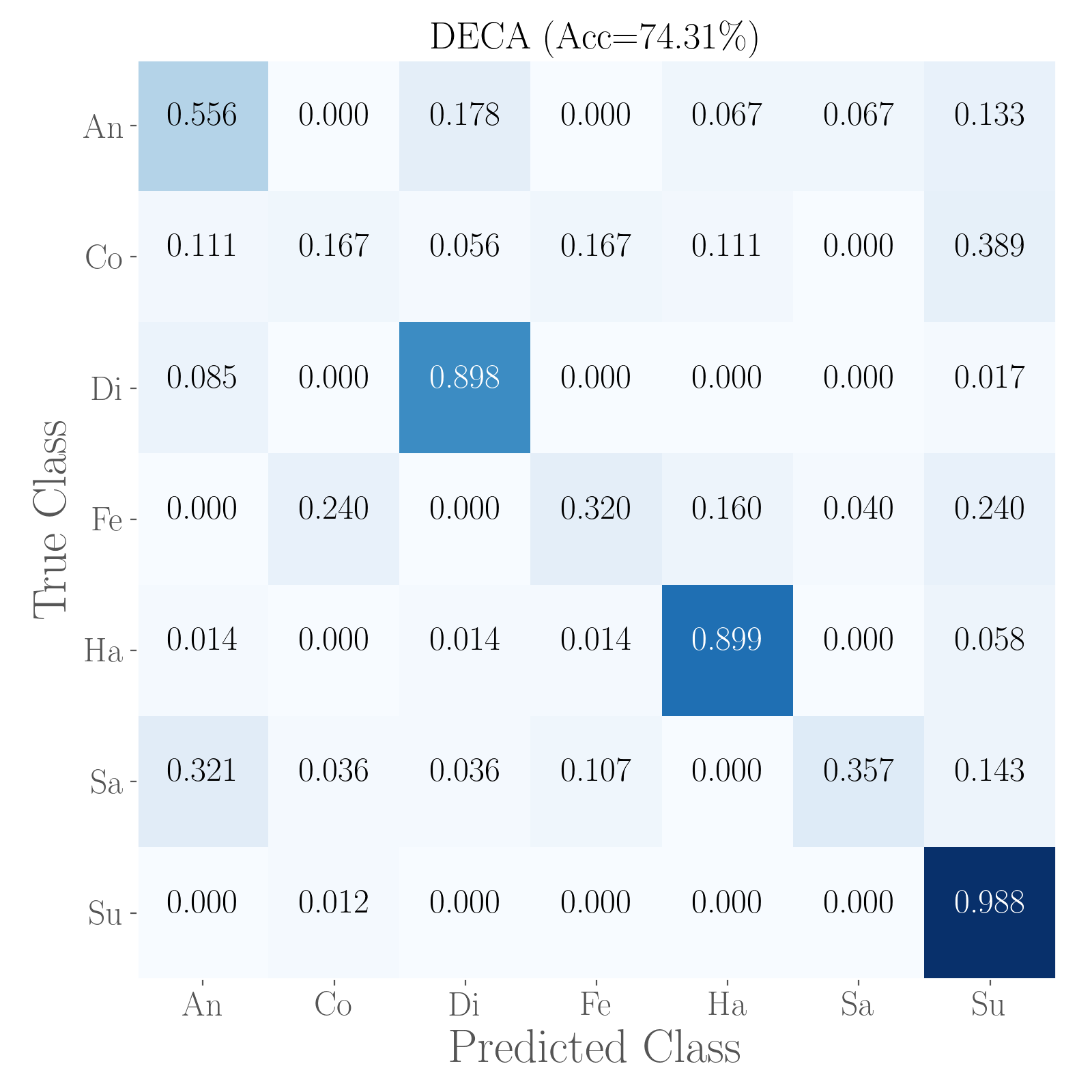}
     \end{subfigure}
     \begin{subfigure}[]{0.41\textwidth}
         \centering
         \includegraphics[width=\textwidth]{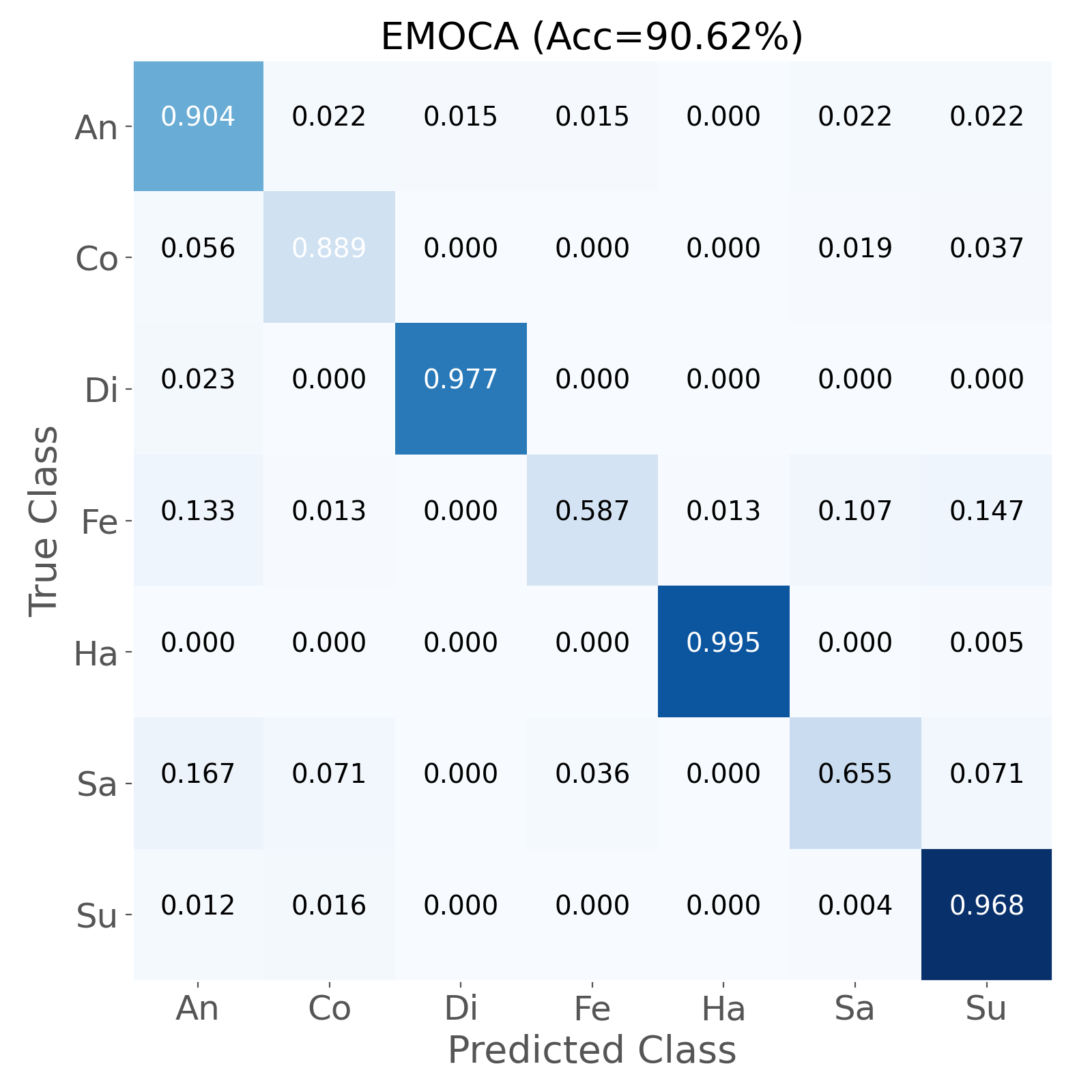}
     \end{subfigure}     
    \caption{Discrete Emotion Recognition Results on \textbf{CK+ Dataset}.}
    \label{fig:disc-emo-ckplus-results}
\end{figure}

\noindent \textbf{Discrete Emotion Recognition.} \Cref{fig:disc-emo-cfee-results} and \Cref{fig:disc-emo-ckplus-results} compares class-wise performance of all four 3D face models on CFEE and CK+ datasets respectively. While all the 3D models achieved reasonably good classification accuracy, EMOCA demonstrates the best performance in terms of mean accuracy on both the datasets. The remaining 3D face models perform consistently well on the positive emotions, i.e., the happy and surprise classes, but for negative emotions (angry, sad, fear, disgust), they have relatively poor performance in most of the cases. t-SNE distributions of the 3D face expressions features illustrated in \Cref{fig:tsne} show similar trends. 

\end{appendices}

\end{document}